\documentclass[lettersize,journal]{IEEEtran}
\usepackage{amsmath,amsfonts}
\usepackage{algorithmic}
\usepackage{array}
\usepackage[caption=false,font=normalsize,labelfont=sf,textfont=sf]{subfig}
\usepackage{textcomp}
\usepackage{stfloats}
\usepackage{url}
\usepackage{verbatim}
\usepackage{graphicx}
\def\BibTeX{{\rm B\kern-.05em{\sc i\kern-.025em b}\kern-.08em
    T\kern-.1667em\lower.7ex\hbox{E}\kern-.125emX}}
\usepackage{balance}
\usepackage{multirow}
\usepackage{algorithm}
\usepackage{graphicx}
\usepackage{cite}
\usepackage{enumitem}
\usepackage{hyperref}

\usepackage{color,soul}

\begin{document}
\title{Unicorn: A Universal and Collaborative Reinforcement Learning Approach Towards Generalizable Network-Wide Traffic Signal Control}
\author{Yifeng Zhang, Yilin Liu, Ping Gong, Peizhuo Li, Mingfeng Fan, Guillaume Sartoretti
\thanks{This work is supported by A*STAR, CISCO Systems (USA) Pte. Ltd and National University of Singapore under its Cisco-NUS Accelerated Digital Economy Corporate Laboratory (Award I21001E0002) (\textit{Corresponding author: Mingfeng Fan}).}
\thanks{Yifeng Zhang, Ping Gong, Peizhuo Li, Mingfeng Fan, and Guillaume Sartoretti are with the Department of Mechanical Engineering, National University of Singapore (E-mail: \{yifeng, e1133090, e0376963\}@u.nus.edu, \{ming.fan, guillaume.sartoretti\}@nus.edu.sg).}
\thanks{Yilin Liu is with the State Key Laboratory of Networking and Switching Technology, Beijing University of Posts and
Telecommunications (E-mail: liuyilin10@bupt.edu.cn).}
}

\maketitle

\begin{abstract}
\label{abstract}
Adaptive traffic signal control (ATSC) is crucial in reducing congestion, maximizing throughput, and improving mobility in rapidly growing urban areas. 
Recent advancements in parameter-sharing multi-agent reinforcement learning (MARL) have greatly enhanced the scalable and adaptive optimization of complex, dynamic flows in large-scale homogeneous networks.
However, the inherent heterogeneity of real-world traffic networks, with their varied intersection topologies and interaction dynamics, poses substantial challenges to achieving scalable and effective ATSC across different traffic scenarios.
To address these challenges, we present Unicorn, a universal and collaborative MARL framework designed for efficient and adaptable network-wide ATSC.
Specifically, we first propose a unified approach to map the states and actions of intersections with varying topologies into a common structure based on traffic movements.
Next, we design a Universal Traffic Representation (UTR) module with a decoder-only network for general feature extraction, enhancing the model’s adaptability to diverse traffic scenarios.
Additionally, we incorporate an Intersection Specifics Representation (ISR) module, designed to identify key latent vectors that represent the unique intersection's topology and traffic dynamics through variational inference techniques. 
To further refine these latent representations, we employ a contrastive learning approach in a self-supervised manner, which enables better differentiation of intersection-specific features.
Moreover, we integrate the state-action dependencies of neighboring agents into policy optimization, which effectively captures dynamic agent interactions and facilitates efficient regional collaboration.
Through comprehensive evaluations against other advanced ATSC methods across eight different traffic datasets, 
our empirical findings reveal that Unicorn consistently outperforms other methods across various evaluation metrics, highlighting its superiority and adaptability in optimizing traffic flows in complex, dynamic traffic networks.
The code is available at \url{https://github.com/marmotlab/Unicorn}.

\end{abstract}

\begin{IEEEkeywords}
Generalizable adaptive traffic signal control, multi-agent reinforcement learning, contrastive learning

\end{IEEEkeywords}

\section{Introduction}
\label{introduction}
Due to rapid urbanization and growing traffic demands, traffic congestion remains a critical challenge in ever-expanding urban areas, often causing significant economic and societal issues~\cite{wei2019survey, haydari2020deep}. 
According to the INRIX 2023 Global Traffic Scorecard~\cite{INRIX2023}, New York City ranked as the most congested urban area in 2023, with drivers losing an average of 101 hours to traffic jams, translating to an economic loss of over \$9.1 billion in wasted time.
Enhancing the performance of existing traffic signal control (TSC) systems without modifying infrastructure (e.g., road layouts, road capacities) is one of the most practical and effective ways to alleviate congestion, as most systems still rely on fixed-time control, which struggles to manage dynamic and complex urban traffic flows. 
This highlights the effectiveness and importance of adaptive traffic signal control (ATSC) in addressing urban congestion.
However, widely deployed ATSC systems such as SCOOT~\cite{hunt1982scoot} and SCATS~\cite{pr1992scats} are mainly rule-based or rely heavily on historical data, limiting their ability to respond effectively to complex, dynamic, real-time traffic conditions.

\begin{figure*}[t!]
    \centering
    \includegraphics[width=0.95\linewidth]{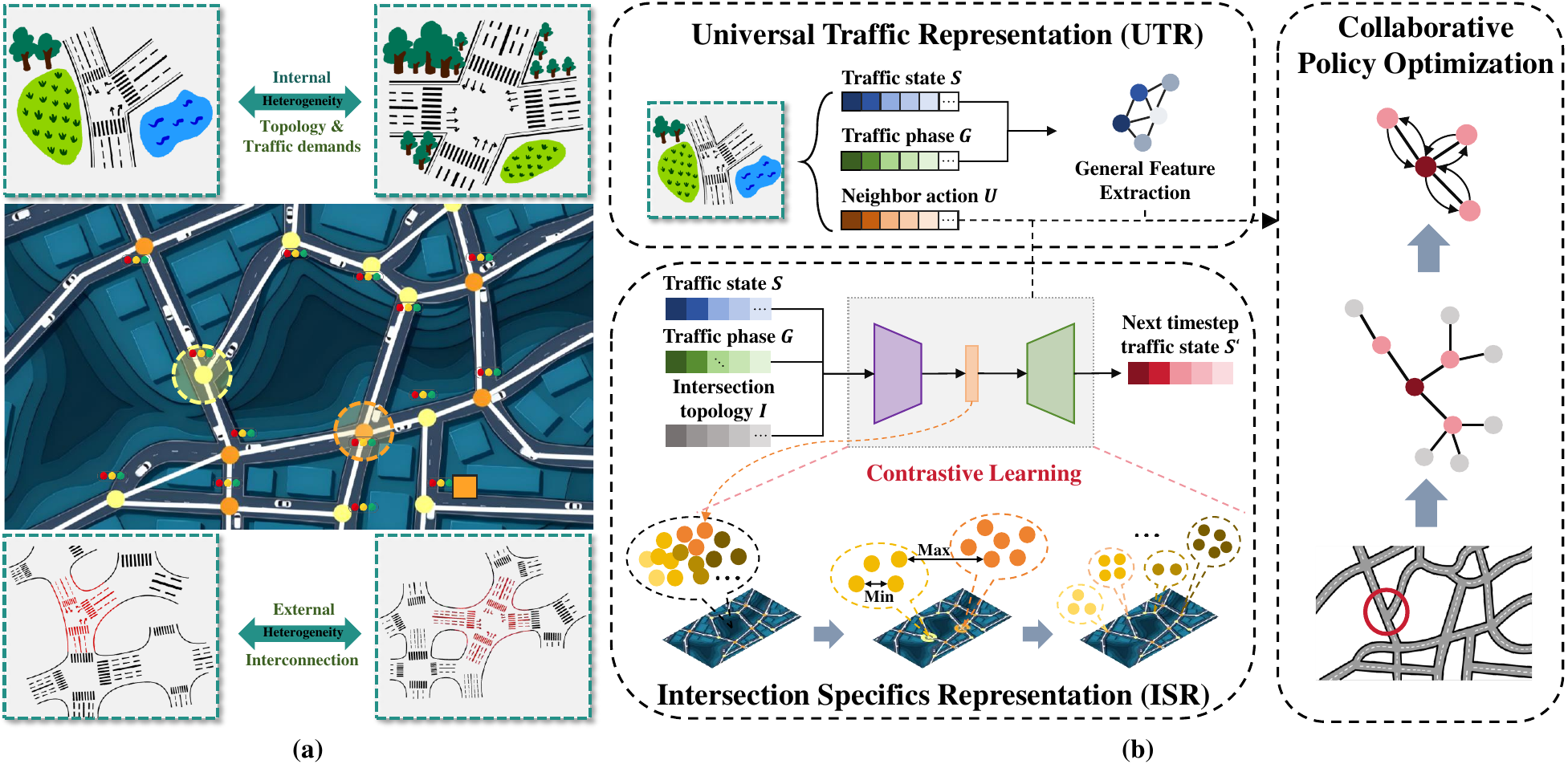}
    \centering
    \vspace{-0.2 cm}
    \caption{
    (a) illustrates the two main sources of heterogeneity in complex real-world traffic networks with multiple intersections: heterogeneity in intersection topology structures and traffic demands (internal), and heterogeneity in interconnection relationships between connected intersections (external).
    (b) provides an overview of our proposed Unicorn framework, which consists of three key components: (1) a Universal Traffic Representation (UTR) module for unified state-action representation and general feature extraction across intersections with diverse topology structures, (2) an Intersection Specifics Representation (ISR) module to enhance the learning of diverse intersection-specific features within the traffic network, and (3) a Collaborative Learning algorithm that strengthens neighborhood coordination and collaboration by leveraging unified state-action dependencies with neighboring intersections.
    }
    \vspace{-0.6cm}
    \label{fig:unicorn_framework}
\end{figure*}

In recent years, data-driven multi-agent reinforcement learning (MARL), particularly decentralized MARL, has shown promising progress in optimizing large-scale and complex traffic flows.
Existing MARL-based ATSC methods can be broadly categorized into two main directions: multi-agent coordination and cooperation, and cross-scenario generalization.
In the former, methods have utilized advanced graph neural networks (GNNs) to aggregate spatiotemporal information from neighboring agents for implicit coordination~\cite{wei2019colight, wang2020stmarl, yang2021ihg, wang2022meta, lin2023temporal, chen2024learning}, attention mechanisms to capture critical interactions among agents~\cite{zhang2023leveraging, zhang2022multi, ruan2024coslight}, and centralized training and decentralized execution (CTDE) learning frameworks with tailored algorithms to improve the efficiency of agent collaboration~\cite{goel2023sociallight, liu2023gplight, zhou2024cooperative}.
For the latter, recent studies have proposed different universal control frameworks capable of using a single model to manage heterogeneous intersections with varied topologies~\cite{zheng2019learning, vaswani2017attention, liang2022oam, jiang2024general, zhang2024HeteroLight}, as well as efficient meta-learning approaches to develop policies adaptable to diverse traffic scenarios~\cite{zang2020metalight, zhu2023metavim, lu2024dualight, jiang2024x}.
Despite these advancements, few studies have successfully integrated universal policy learning and collaborative learning for developing diverse and cooperative control strategies tailored to complex real-world traffic networks.
These traffic networks typically exhibit two key aspects of heterogeneity: the internal heterogeneity of intersection topologies and traffic demands (i.e., local traffic conditions), and the external heterogeneity of interconnection relationships between intersections, as illustrated in Fig.~\ref{fig:unicorn_framework}(a). 
Addressing these challenges is critical for designing generalizable and scalable ATSC frameworks for diverse real-world scenarios.

To address these challenges, we introduce \textbf{Unicorn}, a universal and collaborative MARL framework, designed to achieve diverse and cooperative ATSC across complex, realistic traffic networks. 
An overview of the framework is provided in Fig.~\ref{fig:unicorn_framework}(b).
Specifically, we propose a unified state-action representation approach, which uses the fine-grained structural composition --- traffic movements of intersections as a foundational “common language.”
By doing so, it tightly connects the overall intersection state (real-time traffic conditions of all movements) with phase information (which movements are controlled), enabling a unified representation of intersections with varying topologies and state-action dimensions without being constrained by their structures.
Additionally, we develop a Universal Traffic Representation (UTR) module using a decoder-only network with a cross-attention mechanism to extract crucial phase-specific state representations.
We also apply padding and masking mechanisms to handle variable input-output lengths and ensure effective feature extraction across different intersections and traffic networks.

Furthermore, we present an Intersection Specifics Representation (ISR) module, which integrates a Variational Autoencoder (VAE) designed to generate intersection-specific latent vectors through variational inference techniques.
In particular, our ISE module takes the combination of the current traffic state vector, phase state vector, and intersection topology vector as inputs, to reconstruct predictions for the next traffic state. 
This enables agents to integrate unique intersection topology details, along with the traffic flows and transition dynamics into a compact latent space. 
To enhance representation capabilities, we design a contrastive-learning-based approach that aligns latent vectors from the same agent while separating those from different agents in the latent space. 
This targeted design preserves intersection-specific consistency while structuring diversity in the latent space, allowing the model to learn more effectively by organizing and distinguishing the varied characteristics of different intersections within complex and dynamic traffic networks.
Lastly, we combine the phase features from the two modules to generate both the policy and the value. 
In particular, we incorporate privileged information on neighboring actions into the value estimation and introduce an action cross-attention mechanism to capture the complex and dynamic interactions within the local neighborhood.
Unlike the common practice of using attention to aggregate phase or neighboring traffic state features, our design effectively captures the state–action dependencies at intersections, thereby enhancing value estimation and enabling more efficient and stable collaborative policy learning.
Moreover, we construct a unified neighbor action vector that aligns heterogeneous action semantics across agents, enabling the model to learn generalizable collaborative strategies in diverse traffic networks.

To validate the effectiveness of the Unicorn framework, we conduct extensive experiments across eight traffic datasets, comparing our approach to both traditional ATSC and advanced MARL methods.
These datasets~\cite{chu2019multi, ault2021reinforcement, jiang2024general} include three homogeneous synthetic networks and five heterogeneous real-world networks.
These empirical results demonstrate that our method significantly outperforms other advanced MARL approaches in various key traffic metrics. 
Moreover, the joint training results, where a single model is trained simultaneously on multiple traffic datasets, show that our approach performs comparably to, or better than models trained separately on individual datasets, while consistently surpassing other state-of-the-art universal TSC learning algorithms.
Specifically, we show that our UTR module effectively handles complex intersections with varied topologies, while our ISR module enhances the representation capacity of parameter-sharing models for varied scenarios. 
This improved representation facilitates downstream collaborative strategy learning, leading to superior control performance. 
These findings highlight the effectiveness and robustness of the Unicorn framework in optimizing traffic flows in complex, real-world traffic networks.

This work builds upon our previous study that focused on heterogeneous TSC~\cite{zhang2024HeteroLight}, which introduced two learning modules for extracting general and intersection-specific features from diverse intersections. 
These features were fused for shared policy learning, and the approach was validated on a single real-world traffic dataset.
Beyond these two preliminary feature extraction modules, we propose several key advancements to develop a more universal and collaborative framework for generalizable network-wide TSC.
In summary, the main contributions of this paper are as follows:
\begin{enumerate}
    \item We propose Unicorn, a novel universal and collaborative MARL framework designed for generalizable ATSC. 
    Unicorn introduces a novel approach to unifying state and action representations based on traffic movements, which directly links the input states and output actions in TSC tasks.
    This design ensures universal and consistent representation across intersections, regardless of their topology or phase configurations. 
    Additionally, it incorporates a UTR module for efficient general feature extraction.
    \item
    We introduce an ISR module that leverages latent variables to capture unique intersection topology and traffic dynamics through variational inference techniques. 
    To further enhance the model’s representation capability, we incorporate self-supervised contrastive learning, which promotes distinguish intersection-specific features and improves the distinction between intersections.
    \item Additionally, we incorporate unified privileged information of neighboring agents and employ attention mechanisms to capture inter-agent interactions, facilitating efficient collaboration within neighborhoods.
    \item We validate our approach on eight traffic datasets, which consists of three synthetic homogeneous networks and five realistic heterogeneous traffic networks, and show its superior performance and robust universality compared to other advanced ATSC methods under both single scenario training and joint multiple scenario training setup.
\end{enumerate}

The structure of this paper is as follows: Section~\ref{related_works} provides a review of related work on adaptive traffic signal control. 
Section~\ref{background} introduces key traffic terminology, MARL principles, and the RL formulation. 
The Unicorn framework is described in detail in Section~\ref{method_Unicorn}. Simulation experiments and results are presented in Section~\ref{experiments}. 
Finally, Section~\ref{conclusion} summarizes the findings and discusses potential future research directions.

\section{Related Work}
\label{related_works}
\subsection{Traditional Traffic Signal Control}
Traditional TSC methods can be broadly classified into \textbf{fixed-time control} and \textbf{adaptive control}.
As discussed by Roess et al.~\cite{roess2004traffic}, fixed-time control operates based on a predetermined phase cycle and timing of phases, yet it struggles to adapt to complex, changing traffic flows.
On the other hand, adaptive control systems, such as SCOOT~\cite{hunt1982scoot} and SCATS~\cite{pr1992scats} adjust signal plans in response to real-time traffic conditions, leveraging data collected from Induction Loop Detectors (ILDs) for more responsive management. 
Furthermore, the max-pressure control~\cite{varaiya2013max} optimizes traffic flow at intersections by minimizing the difference in the vehicle counts between upstream and downstream roads.
However, due to their reliance on hand-crafted rules or simplified models with fixed assumptions, these methods struggle to adapt to dynamic and complex traffic flows, prompting the development of more advanced approaches such as model predictive control (MPC) and schedule-driven methods.
MPC methods~\cite{ye2015two, wu2018distributed} plan a sequence of signal actions based on traffic predictions, which offers flexibility to handle irregular flows. 
However, the high computational cost of real-time optimization, especially in centralized settings, limits their scalability and timely responsiveness in large-scale networks.
Schedule-driven methods~\cite{xie2012schedule, hu2019softpressure, goldstein2018expressive} address this issue by formulating ATSC as a scheduling problem, where vehicles are treated as jobs. 
They achieve scalability by combining computationally efficient scheduling algorithms with lightweight decentralized coordination mechanisms. 
However, as scheduling decisions rely on accurate vehicle arrival predictions, potential errors from sensor noise and traffic fluctuations can lead to suboptimal or unstable control in real-world scenarios.
In a recent study, Wang et al.~\cite{wang2024real} proposed an explicit multi-agent coordination (EMC)-based online planning method for real-time network-level ATSC, focusing on further improving coordination performance, robustness, and scalability.

\subsection{Learning-based Traffic Signal Control}

Recently, MARL has emerged as a promising approach for ATSC, demonstrating great potential in improving network-wide traffic flow.
Previous research mainly focused on two key aspects: \textbf{multi-agent coordination and cooperation}, and \textbf{cross-scenario generalization}.
In the context of multi-agent coordination and cooperation, the majority of existing studies~\cite{wei2018intellilight, wei2019presslight, wei2019colight, chu2019multi, yang2021ihg, zhang2022expression, zhang2022neighborhood, goel2023sociallight, liu2023gplight, zhangCoordLight} have concentrated on the development of advanced decentralized MARL frameworks mainly for homogeneous traffic networks, enabling agent coordination and cooperation either implicitly by optimizing their own objectives without directly considering others' rewards, or explicitly by incorporating other agents' rewards into their optimization objectives.
For instance, IntelliLight~\cite{wei2018intellilight}, PressLight~\cite{wei2019presslight}, MPLight~\cite{chen2020toward}, and Advanced-XLight~\cite{zhang2022expression} enhanced state and reward spaces by leveraging refined traffic information, such as image-like states, pressure or advanced traffic state, to achieve implicit coordination.
Alternatively, approaches like CoLight~\cite{wei2019colight}, STMARL~\cite{wang2020stmarl}, IHG-MA~\cite{yang2021ihg}, MetaGAT~\cite{wang2022meta}, GPLight~\cite{liu2023gplight}, TeDA-GCRL~\cite{lin2023temporal}, and CoevoMARL~\cite{chen2024learning} employed graph neural networks (GNNs) to facilitate efficient spatio-temporal feature extraction among connected agents, thereby improving representation learning and fostering better coordination.
In contrast to these implicit approaches, explicit methods have focused on directly optimizing multi-agent cooperation~\cite{zhang2022neighborhood, goel2023sociallight, ruan2024coslight, zhou2024cooperative}. 
For example, NC-HDQN~\cite{zhang2022neighborhood} leveraged the correlations between adjacent agents to adjust their observations and rewards, thus promoting neighborhood cooperation.
SocialLight~\cite{goel2023sociallight} presented a decentralized MARL algorithm with a refined counterfactual advantage calculation for scalable cooperation.
More recently, CoordLight~\cite{zhangCoordLight} introduced QDSE, a novel state representation, as well as NAPO, a neighbor-aware MARL algorithm, to improve local decision-making and neighborhood coordination, enabling efficient and scalable network-wide traffic optimization.
Additionally, Zhou et al.~\cite{zhou2024cooperative} proposed the MICDRL framework, advancing the CTDE paradigm for cooperative ATSC with an incentive communication mechanism for effective message exchange and a mutual information-based module to model interactions.
Despite their effectiveness, most of these methods focus on homogeneous road networks and lack the ability to address the heterogeneity encountered in realistic networks, limiting their practicality and generalizability to real-world TSC applications.

In addition to this, another important aspect of learning-based ATSC research is cross-scenario generalization, which can be approached in two main ways. 
One direction focuses on designing universal parameter-sharing frameworks to handle diverse traffic networks, allowing a single model to manage multiple intersections with different road structures and phase settings~\cite{zheng2019learning, chen2020toward, oroojlooy2020attendlight, liang2022oam, jiang2024general, wang2024unitsa, zhang2024HeteroLight}. 
This is because previous independent learning approaches, like IA2C and MA2C~\cite{chu2019multi} were developed to allow each agent to learn its own policy,  yet they often lead to suboptimal outcomes due to environmental instability.
To address this issue, Zheng et al.~\cite{zheng2019learning} introduced FRAP, a parameter-sharing method that leverages phase competition, tailored for intersections with diverse topologies and traffic patterns. 
Furthermore, GESA~\cite{jiang2024general} enhanced FRAP by standardizing all intersections into a unified four-way road structure and ensuring consistent state and action spaces for streamlined input and output.
AttendLight~\cite{oroojlooy2020attendlight} introduced an attention-based learning framework for universal policy learning applicable to any intersection configuration. 
Similarly, Liang et al.~\cite{liang2022oam} introduced the Option-Action (OAM) framework, which reformulates phase selection as constrained combinations of lane-level options and designs a universal network that can adapt to intersections with arbitrary phase definitions. 
While these methods achieve universal policy learning across intersections by standardizing topological structures and phase configurations to maintain consistent input and output spaces, they typically rely on a single parameter-shared model with limited representational capacity. 
As a result, such models often fail to capture intersection-specific characteristics, which makes it difficult to achieve adaptability and fine-grained control in diverse and complex traffic scenarios.

Another line of research explores improving models' ability to generalize across scenarios, enabling them to transfer knowledge effectively and adapt to unseen traffic conditions~\cite{zang2020metalight, lu2024dualight, jiang2024x, jiang2024general}.
For instance, MetaLight~\cite{zang2020metalight} introduced a meta-reinforcement learning approach designed for training across multiple scenarios.
DuaLight~\cite{lu2024dualight} introduced a dual-module framework that leverages scenario-specific experiential weights and scenario-shared co-training to enhance decision-making by adapting to intersection-level features within a single scenario while learning generalizable dynamics across multiple scenarios.
Additionally, MetaVIM~\cite{zhu2023metavim} improved policy generalization across varying neighborhood sizes by integrating latent variables and introduced an intrinsic reward to stabilize training in dynamic traffic conditions. 
Despite these advancements, few parameter-sharing methods consider both universal policy diversity and effective, scalable multi-agent collaboration, yet these aspects are crucial for optimizing complex, dynamic real-world traffic flows.

\section{Background}
\label{background}

In this section, we begin by introducing fundamental traffic terminology and then formulate the MATSC problem within a MARL framework. Finally, we describe the RL agent design, which serves as the foundation of our proposed approach.

\subsection{Preliminary}

In this paper, we investigate generalizable network-wide ATSC, which focuses on both heterogeneous networks (where intersections have diverse topology structures) and homogeneous networks (where all intersections share identical topology structures). Typically, a traffic network is determined by four core concepts: incoming and outgoing lanes, traffic movements and movement status, traffic signal phases, and traffic agents within the network, detailed as follows:
\begin{enumerate}[left=0pt, label=\textbullet]
    \item \textbf{Incoming and outgoing lanes}: Incoming lanes are the segments of the road that guide vehicles toward an intersection, while outgoing lanes lead vehicles away. Each road at the intersection comprises multiple lanes; $\mathcal{L}_{in}$ denotes the set of incoming lanes, and $\mathcal{L}_{out}$ denotes the set of outgoing lanes.
    \item \textbf{Traffic movements and movement status}: A traffic movement refers to the connection between an incoming lane ($l_{in}$) to an outgoing lane ($l_{out}$), forming a route for vehicles through the intersection. This movement is denoted as $m(l_{in}, l_{out})$, though $(l_{in}, l_{out})$ is omitted in subsequent content for simplicity. Besides, $A^m$ represents the movement status: if vehicles are permitted to move, the movement is active ($A^m=1$); otherwise, it is inactive ($A^m=0$).
    \item \textbf{Traffic signal phases}:
    Traffic signal phases manage movements at intersections by grouping non-conflicting movements that can proceed simultaneously.
    For a given intersection $i$, the movement set $\mathcal{M}_{i}$ is the union of all movements across phases: $\mathcal{M}_{i} = \bigcup_{p \in \mathcal{P}} M_{i}^{p}$, where $\mathcal{P}$ is the set of all phases, and $M_{i}^{p}$ is the set of active movements in phase $p$.
    \item \textbf{Traffic agents within the network}:
    Traffic agents regulate the flow at intersections by controlling traffic signal phases and their timings. The traffic network is modeled as a multi-agent system, represented by $\mathcal{G}(\mathcal{V}, \mathcal{E})$, where $\mathcal{V}=\{1,\ldots, N\}$ denotes the set of traffic agents, and  $\mathcal{E}$ represents the set of edges connecting intersections through roads. These edges facilitate communication between each agent and its directly connected neighboring agents.
    
\end{enumerate}

\begin{figure*}[t!]
    \centering
    \includegraphics[width=\linewidth]{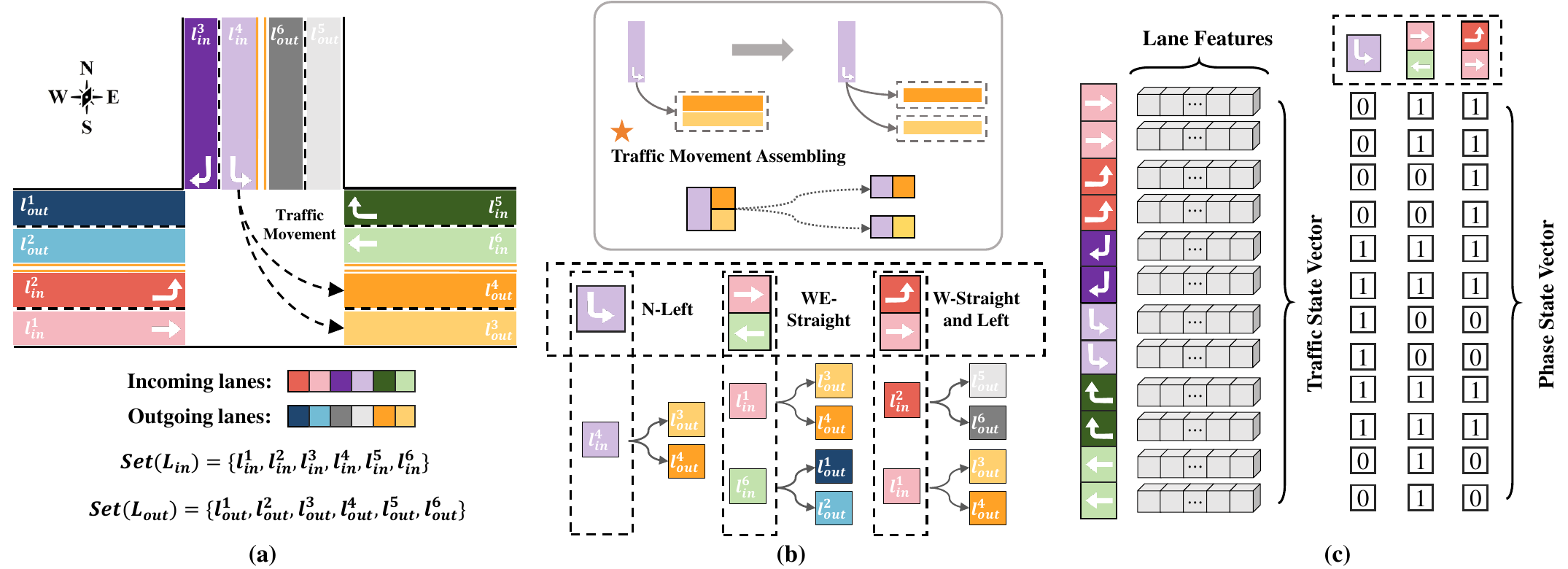}
    \centering
    \vspace{-0.5cm}
    \caption{
    (a) Overview of a typical 3-arm intersection, which consists of six incoming lanes, six outgoing lanes, 12 traffic movements, and three traffic phases.
    (b) The top portion illustrates how each traffic movement is assembled by linking an incoming lane with an outgoing lane. The bottom part shows that each phase consists of a group of activated, non-conflicting traffic movements, thereby establishing the relationship between phases and movements.
    (c) Illustration of the proposed traffic state vector and traffic phase vector (comprising multiple phase state vectors), both constructed based on ordered traffic movements.
    }
    \vspace{-0.6cm}
    \label{fig:intersection}
\end{figure*}

As an example, Fig.~\ref{fig:intersection}(a) illustrates a three-armed intersection with six incoming lanes and six outgoing lanes (two lanes per road), allowing for 12 possible traffic movements, assuming each incoming lane connects to multiple outgoing lanes. 
The intersection operates under three traffic signal phases, as illustrated in Fig.~\ref{fig:intersection}(b), which shows how traffic movements are assembled and indicates the active traffic movements corresponding to each traffic phase.

\subsection{MARL framework for MATSC}

We formulate the MATSC problem as a MARL task, given that each intersection is managed by an individual RL agent in a fully decentralized framework. 
Specifically, it is formalized as a Decentralized Partially Observable Markov Decision Process (Dec-POMDP) \cite{oliehoek2016concise}, defined by the tuple $\langle\mathcal{V}, \mathcal{S}, \mathcal{A}, \mathcal{O}, \mathbb{P}, \mathcal{R}, \gamma, \rho_0 \rangle$ where $\mathcal{V}$ denotes the set of learning agents within the network, and $s \in \mathcal{S}$ represents the global traffic state, which individual agents cannot directly observe.
At each time step $t$, an agent $i \in \mathcal{V}$ receives a local observation $o_i^t$ via the observation function $\mathcal{O}(s^t, i)$ and selects an action $a_i^t \in \mathcal{A}_i$ according to its policy $\pi(\cdot \,|\, o_i^t)$. 
The joint action $\mathbf{a^t}\in \mathcal{A}$, formed by all agents, determines the next state $s_{t+1}$ via the transition function $\mathbb{P}(s^{t+1}| s^t, \mathbf{a^t})$. 
Each agent is rewarded with $r_i^t$, derived from its reward function $\mathcal{R}_i(o_i^t, a_i^t)$. 
The parameters $\gamma$ and $\rho_0$ denote the discount factor and the initial state distribution, respectively. 
The MATSC system aims to find an optimal joint policy $\pi^*$ that maximizes the expected cumulative discounted rewards for all agents:
$\pi^* = \underset{\pi}{\operatorname{argmax}} \; \mathbb{E}_{\tau}\left[\sum_{i=1}^{N} \sum_{t=1}^{\infty} \gamma^t r_i^t\right],$
where $\tau = {(\mathbf{o}^t, \mathbf{a}^t, \mathbf{r}^t)}_{t=1}^{t_e}$ represents the global trajectory of length $t_e$. 
This formulation allows agents to optimize individual rewards while collaboratively enhancing traffic flow, providing a scalable solution for improving overall network performance.

\subsection{RL Agent Design}

In this subsection, we detail the formulation of the RL agent, including the design of its state, action, and reward.

\subsubsection{State}
\label{state}

At each time step, the RL agent receives quantitative descriptions of the environment as a state to decide an action. 
Commonly used states are divided into two types: image-like and lane-feature states. 
Image-like states capture traffic dynamics through grid-based visual representations~\cite{garg2018deep, wei2018intellilight, ma2021deep}, offering rich information but requiring substantial computational resources and being more sensitive to noise. 
Conversely, lane-feature states are simple yet efficient and aggregate lane-specific data such as queue length, vehicle count, average speed, and pressure~\cite{chu2019multi, wei2019colight, chen2020toward, zhang2022expression, oroojlooy2020attendlight, goel2023sociallight, zhang2022multi}, to represent local traffic conditions. 
Moreover, these lane features can be readily obtained from induction loop detectors (ILDs) or cameras, making them highly applicable to real-world scenarios.
In this work, we utilize a widely adopted combination of multiple lane features for the state definition~\cite{jiang2024general}. 
Unlike existing approaches, we base these features on traffic movements to capture traffic conditions more accurately, enabling a more precise depiction of states across intersections with diverse topologies, as illustrated in Fig.\ref{fig:intersection}(c). 
Further details are provided in Sec.~\ref{method_state}.

\subsubsection{Action}

Following prior RL-based ATSC studies~\cite{wei2019presslight, wei2019colight, chen2020toward, chu2019multi, goel2023sociallight} and to ensure fair comparisons, we define each agent’s action space as a finite set of collision-free traffic signal phases. At each decision step, agents select and apply a phase from this set for a fixed duration, without being constrained to a preset cycle. Moreover, our approach is versatile, supporting alternative action spaces, such as maintaining the current phase or transitioning to the next phase within a predefined cycle and duration limits~\cite{mannion2016experimental, wei2018intellilight, wang2024traffic}, ensuring compatibility with realistic traffic signal control scenarios.

\subsubsection{Reward}

The reward for each traffic agent is defined as the negative sum of queue lengths at the intersection:
$R(t)=-\left(\sum_{l_{in} \in \mathcal{L}_{in}} Q^{l_{in}} + \sum_{l_{out} \in \mathcal{L}_{out}} Q^{l_{out}}\right)$, where $Q^{l_{in}}$ and $Q^{l_{out}}$ are the queue lengths on the incoming lane $l_{in}$ and outgoing lane $l_{out}$, respectively. 
Queue lengths are measured using lane-area ILDs with a detection range of 50 meters, installed on both incoming and outgoing lanes near the intersection.
This reward design couples the rewards of neighboring agents, encouraging improved traffic flow within the local neighborhood and fostering collaboration among agents.

\section{Unicorn}
\label{method_Unicorn}

In this section, we present Unicorn, a universal and collaborative MARL framework designed for efficient and adaptable network-wide ATSC.
We first construct a unified representation by mapping the states and actions of intersections with diverse topologies and phase configurations into a common structure based on traffic movements, as illustrated in Fig.~\ref{fig:intersection}.
Building on this, we design two key modules: the UTR module and the ISR module, both shown in Fig.~\ref{fig:method}.
The UTR module uses a decoder-only network to extract generalizable features, while the ISR module applies variational inference to learn latent features that capture each intersection's unique topological structure and traffic dynamics.
To further enhance feature distinctiveness, we employ a self-supervised contrastive learning strategy to refine the latent features learned by the ISR module.
The features produced by the UTR and ISR modules are then integrated to generate the final policy and value functions. 
Finally, we introduce a decentralized collaborative MARL algorithm that integrates key dependencies from neighboring agents through the same unified state-action representation into the policy optimization process, enabling generalizable coordination across heterogeneous neighborhoods, as also shown in Fig.~\ref{fig:method}.
In summary, Unicorn employs a unified state–action representation as a bridge, enabling a modular network architecture to address internal heterogeneity by allowing a single parameter-sharing model to handle diverse intersections, and uses an attention mechanism in collaborative learning to tackle external heterogeneity by capturing interdependencies among neighbors regardless of their count or interconnections. 

\subsection{Unified State-Action Representation}
\label{method_state}

Establishing a unified representation framework for state and action spaces enables parameter-sharing RL agents to manage intersections with diverse topologies more effectively.
By defining states and actions regarding traffic movements, this framework provides a consistent and interpretable “language” that bridges varying intersection configurations.
This unified representation ensures that traffic conditions (input states) are systematically linked to signal phases (output actions), allowing the model to generalize effectively across heterogeneous intersections and networks. 
In this subsection, we define the traffic state vector, traffic phase vector, intersection topology vector, and neighbor action vector, which together form the foundation for the subsequent learning processes.

The \textbf{traffic state vector} $S_i({t})$ for an intersection $i$ consists of the state vectors $S^{m}_{i}(t)$ ($m \in \mathcal{M}_{i}$) corresponding to all available traffic movements, expressed as: $S_{i}(t) \in \mathbb{R}^{|\mathcal{M}_{i}| \times 8}=[S^{m}_{i}(t) \, | \, m \in \mathcal{M}_{i}]$, where $S^{m}_{i}(t)$ is defined as follows: 
\begin{equation}
\!S^{m}_{i}(t)\! \in \!\mathbb{R}^{8}\! =\! [A^{m},\! Q^{l_{in}},\! Q^{l_{out}},\! N^{l_{in}},\! N^{l_{out}},\! O^{l_{in}},\! O^{l_{out}}, \!C^{l_{out}}].    
\end{equation}
Here, $A^{m}$ represents the activation status of the traffic movement $m$, while $Q^{l_{in}}$ and $Q^{l_{out}}$ denote the number of stopped vehicles (queue length) on the incoming and outgoing lanes associated with movement $m$, respectively. 
Similarly, $N^{l_{in}}$ and $N^{l_{out}}$ indicate the number of moving vehicles on the incoming and outgoing lanes, and $O^{l_{in}}$ and $O^{l_{out}}$ are the occupancy percentages (\%) of the incoming and outgoing lanes, respectively, measured as the proportion of space occupied by vehicles. 
Lastly, $C^{l_{out}}$ is a time-independent vector with binary elements indicating whether the outgoing lane is controlled by a traffic signal, acknowledging that many real-world intersections operate without traffic signals.

The \textbf{traffic phase vector} $G_i$ for an intersection $i$ is defined as a collection of time-invariant \textbf{phase state vectors} ($G^p_i$). 
It is represented as: $G_i \in \mathbb{R}^{|\mathcal{P}_i| \times |\mathcal{M}_i|} = [G^p_i \mid p \, \in \, \mathcal{P}_i]$, where 
$G^p_i$ specifies the activation status of all traffic movements for a given phase $p \in \mathcal{P}_i$ at the intersection:
\begin{equation}
    G^p_i \in \mathbb{R}^{|\mathcal{M}_i|} = [1 \text{ if } m \in \mathcal{M}_{i}^{p} \text{ else} 0 \mid m \in \mathcal{M}_i].
\end{equation}
This traffic phase vector accurately reflects the relationship between phase control and traffic states through element-wise correspondence.
An illustration of the traffic state vector and traffic phase vector for a typical 3-arm intersection is shown in Fig.~\ref{fig:intersection}(c).
Each element $G^p_i$ is then fed into the RL model to generate a phase-specific feature and its corresponding action probability (detailed in Sec.~\ref{method_utr}), ensuring a consistent mapping between input phase states and output actions across diverse intersection topologies and traffic conditions.

The \textbf{intersection topology vector} $I_i$ for the intersection $i$ is a time-invariant vector, which is denoted as: 
\begin{equation}
I_i = [T_{\text{tl}}, L^{\mathrm{in}}, V_{\text{max}}^{in}, N_{\text{l}}^{in}, N_{\text{m}}^{in}, L^{\mathrm{out}}, V_{\text{max}}^{out}, N_{\text{l}}^{out}].
\end{equation}
Here, $T_{\text{tl}}$ is a one-hot vector indicating the type of phase settings at the intersection.
$L^{\mathrm{in}}$ is the average lane length of all incoming roads, while $V_{\text{max}}^{in}$ represents the average maximum speed permitted on these roads. $N_{\text{l}}^{in}$ denotes the total number of lanes on the incoming roads, and $N_{\text{m}}^{in}$ indicates the total number of traffic movements associated with them. Similarly, for outgoing roads: $L^{\mathrm{out}}$, $V_{\text{max}}^{out}$, and $N_{\text{l}}^{out}$ represent the average lane length, maximum speed, and the number of lanes, respectively.
This vector encapsulates the intersection’s layout and traffic regulations, providing a multidimensional representation of its structural and operational characteristics.

Additionally, the unified \textbf{neighbor action vector} $U_i(t) \in \mathbb{R}^{|\mathcal{M}_i|}$ captures the current phase states of directly connected neighboring intersections based on the current activation status of the outgoing lanes (as they are also the incoming lanes of neighboring intersections), which is expressed as:
\begin{equation}
U_i[m]= \begin{cases} 1 & \text{ if } l_{out} \in m \text{ is active under } \bar{p}_{j}, \\ 0 & \text{ otherwise,} \end{cases}
\end{equation}
where $\bar{p}_{j}$ denotes the current phase of the neighboring intersection $j$, and $U_i[m]$ is the value in $U_{i}(t)$ corresponding to the traffic movement $m \in \mathcal{M}_i$.
This vector essentially reflects the current actions of the neighboring intersections.

Furthermore, we employ a padding and masking mechanism to maintain dimensional consistency in the unified representation, facilitating parameter sharing and batch training. 
Specifically, the movement and phase dimensions are zero-padded to a fixed length (determined by network complexity), and binary phase masks are applied to exclude padded elements from attention computation, action probabilities, and value estimates, ensuring that only valid phases are considered. Additional details are provided in our \href{https://www.dropbox.com/scl/fi/54bepi43mm8nkaqn54607/Supplementary_Material-T-ITS-Unicorn.pdf?rlkey=71aeqnpyg9zzyq12tlwjb1808&st=ke46mufi&dl=0}{Supplementary Material}.

\subsection{Universal Traffic Representation (UTR) Module}
\label{method_utr}

The UTR module aims to facilitate shared policy learning for ATSC in real-world traffic networks, especially those with heterogeneous configurations. To achieve this, we design a \emph{general feature extraction} (GFE) network within the module, implemented in a decoder-only architecture. Unlike traditional parameter-sharing methods, our approach overcomes the fixed input-output dimension constraint, enabling efficient extraction of key traffic features across diverse intersection layouts.

\begin{figure*}[t!]
    \centering
    \includegraphics[width=0.95\linewidth]{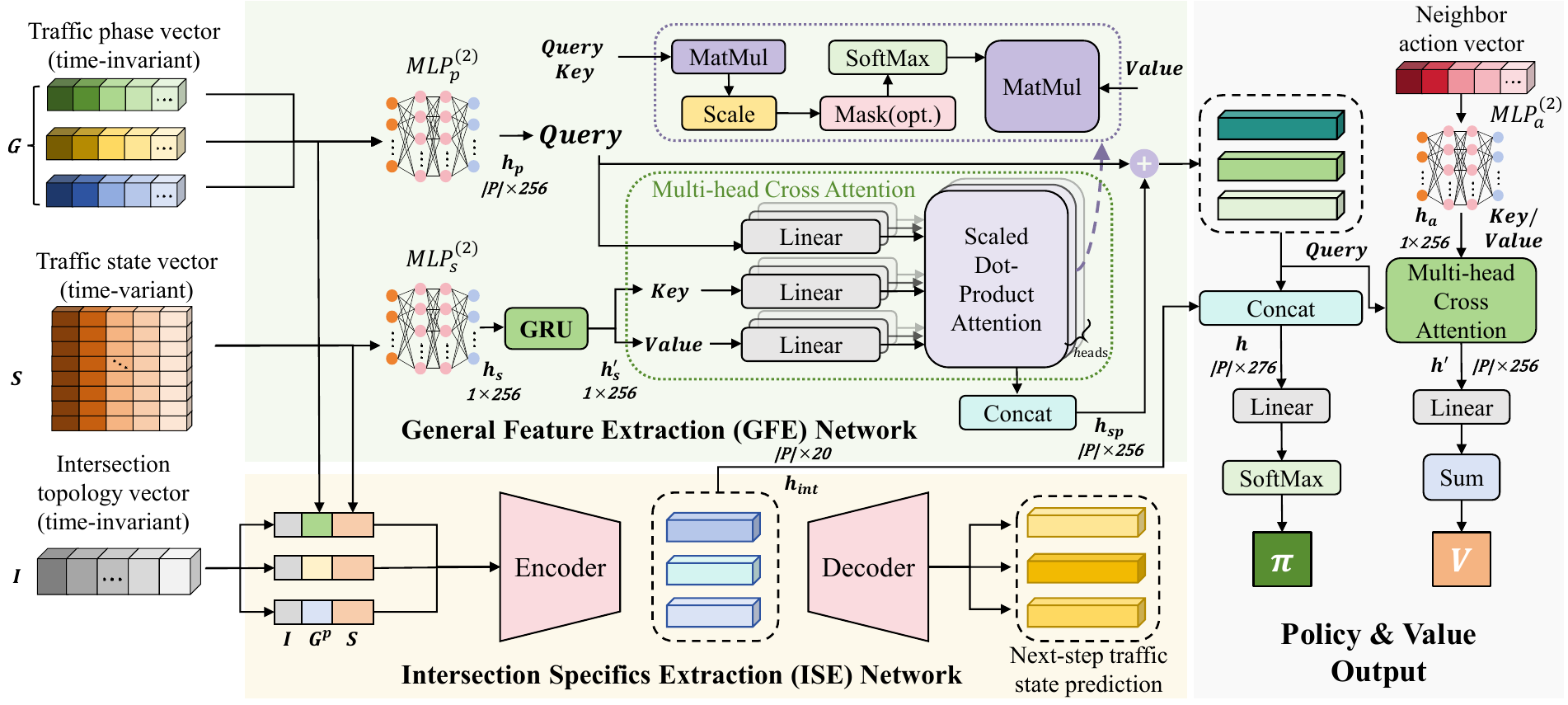}
    \centering
    \caption{
    A detailed illustration of the input vectors (left), the UTR module's General Feature Extraction (GFE) network (top), and ISR module's Intersection-Specific Extraction (ISE) network (bottom), along with the feature integration for policy function and value function outputs (right).
    }
    \vspace{-0.6cm}
    \label{fig:method}
\end{figure*}

The GFE network, as illustrated in Fig.~\ref{fig:method}, first transforms the traffic state vector $S_i^t$ of agent $i$ (representing intersection $i$) at time step $t$ into a high-dimensional feature vector using a multi-layer perceptron (MLP) comprising two linear layers, expressed as:
$\mathbf{h}_{s} \in \mathbb{R}^{d} = \mathbf{MLP}^{(2)}_{s} \left(S_i^t\right),$ where $\mathbf{h}_{s}$ is the resulting state feature vector and $d$ denotes the feature dimension. 
Then, the state feature vector $\mathbf{h}_{s}$ is fed into a Gated Recurrent Unit (GRU)~\cite{chung2014empirical}, which selectively integrates crucial historical local traffic state information while forgetting irrelevant data.
The updated state feature vector $\mathbf{h}^{'}_{s}$ is computed as:
$\mathbf{h}^{'}_{s} \in \mathbb{R}^{d} = \mathbf{GRU} \left(\mathbf{h}_{s}, \mathbf{h}^{GRU}_{(s, t-1)} \right),$
where $\mathbf{h}^{GRU}_{(s, t-1)}$ represents the hidden state produced by the GRU at the previous decision step $t-1$. 
Similarly, the traffic phase vector $G_i$ is transformed into a high-dimensional feature vector using an additional MLP comprising two linear layers. The resulting phase feature vector $\mathbf{h}_{p}$ is calculated as:
$\mathbf{h}_{p} \in \mathbb{R}^{|\mathcal{P}_i| \times d} = \mathbf{MLP}^{(2)}_{p}\left({G}_i\right)$.
To create phase-conditioned state feature vectors, we leverage a multi-head cross-attention mechanism~\cite{vaswani2017attention}. 
Specifically, for each attention head $h$ (with a total of 4 heads), the query vector $\mathbf{Q}_{h}$, derived from the phase feature vector $\mathbf{h}_{p}$, is calculated as $\mathbf{Q}_{h} \in \mathbb{R}^{|\mathcal{P}_i| \times d} = \mathbf{h}_{p} \, W_h^Q$.  
The updated state feature vector $\mathbf{h}^{'}_{s}$, output by the GRU, is used to compute the key vector $\mathbf{K}_{h}$ and the value vector $\mathbf{V}_{h}$: $\mathbf{K}_{h} \in \mathbb{R}^{1 \times d} = \mathbf{h}^{'}_{s} \, W_h^K$ and $\mathbf{V}_{h} \in \mathbb{R}^{1 \times d} = \mathbf{h}^{'}_{s} \, W_h^V$.
Here, $W_h^Q$, $W_h^K$, and $W_h^V$ are learnable parameters.
The attention score for each head is then computed as:
$\operatorname{Att}_h(\mathbf{Q}_h, \mathbf{K}_h, \mathbf{V}_h) = \operatorname{softmax}\left(\frac{\mathbf{Q}_h\left(\mathbf{K}_h\right)^T}{\sqrt{d}}\right) \mathbf{V}_h.$
Finally, the attention scores from all heads are concatenated and passed through a linear transformation, yielding the aggregated feature vector: $\mathbf{h}_{sp} \in \mathbb{R}^{|\mathcal{P}_i| \times d} = \operatorname{Concat}\left(\operatorname{Att}_1, \ldots, \operatorname{Att}_4\right) \, W^O,$
where $W^O$ represents the learnable parameters of the output layer.
The aggregated feature vector $\mathbf{h}_{sp}$ will be further utilized to compute the output policy and value functions.
Hence, the UTR module leverages the specified phase state vector to guide the GFE network in aggregating crucial traffic states for each phase. 
This process ensures that the extracted features are both relevant to the traffic conditions and tailored to the control requirements of each phase at the intersection.

\subsection{Intersection Specifics Representation (ISR) Module}
\label{method_isr}
To enhance the representation capability of our Unicorn framework for diverse traffic scenarios, the ISR module introduces an \emph{Intersection-Specific Extraction} (ISE) network to generate latent vectors that integrate phase state vectors and intersection topology information from multiple agents. Additionally, we incorporate a contrastive learning approach to refine the representation learning of these latent vectors.

\subsubsection{ISE Network}

Although parameter sharing is effective for learning scalable decentralized control strategies in multi-agent systems, its limited model representation capacity often results in suboptimal solutions, particularly in scenarios involving heterogeneous agents. 
To mitigate this issue, we introduce the ISE network, as shown in Fig.~\ref{fig:method}, which generates more accurate low-dimensional representations that capture both the static topology and dynamic traffic patterns across various intersections, thereby enhancing the parameter-sharing mechanism and enabling the learning of more diverse policies. 

The backbone of the ISE network is a VAE~\cite{kingma2013auto}, which takes as input a tuple composed of the current traffic state vector $S_i^t$, the phase state vector $G_i^p$, and the intersection topology vector $I_i$ for agent $i$ at time step $t$ with a given traffic phase $p$, denoted as $x_i^p(t) = [S_i^t, G_i^p, I_i]$.
The VAE employs an encoder, parameterized by $\xi$, to approximate the true posterior distribution $p(z_i^p \,|\, x_i^p)$ using a variational distribution $q_{\xi}(z_i^p \,|\, x_i^p)$. 
Specifically, the encoder maps the input $x_i^p$ to the parameters of a Gaussian distribution, represented by the mean vector $\mu_{i}^{p}$ and the standard deviation vector $\sigma_i^p$:
\begin{equation}
(\mu_i^p, \sigma_i^p) = \textbf{Encoder}_{\xi}(x_i^p).
\end{equation} 
A latent variable $z_i^p$ is sampled from this distribution using the reparameterization trick: $z_i^p = \mu_i^p + \sigma_i^p \,\odot\, \epsilon$, where $\epsilon \sim \mathcal{N}(0, I)$.
The decoder, parameterized by $\phi$, leverages the latent variable $z_i^p$ to reconstruct the predicted next state $S_{i,p}^{t+1}$, described by:
\begin{equation}
S_{i,p}^{t+1} \sim p_\phi\left(S_{i,p}^{t+1} \,|\, z_i^p\right) = \textbf{Decoder}_\phi(z_i^p).
\end{equation}

Notably, multiple predictions are generated, each corresponding to a specific input phase state vector. However, during each training step, only the prediction associated with
the selected phase (denoted as $\bar{p}$) is actively used.
The training objective aims to maximize the Evidence Lower Bound (ELBO)~\cite{kingma2013auto}, denoted as $L_i^{vae}$, which serves as a surrogate for maximizing the log-likelihood $\log p\left(\hat{S}_i^{t+1} \mid x_i^{\bar{p}}\right)$. This log-likelihood represents the probability of observing the true next state $\hat{S}_i^{t+1}$ given $x_i^{\bar{p}}$. The ELBO is computed as:
\begin{equation}
\label{eqn:ELBO}
\mathbb{E}_{q_{\xi}(z_i^{\bar{p}} | x_i^{\bar{p}})}\left[\log p_\phi\left(\hat{S}_i^{t+1} | z_i^{\bar{p}} \right)\right]\! -\!\mathrm{KL}\left[q_{\xi}(z_i^{\bar{p}} | x_i^{\bar{p}}) \| p(z_i^{\bar{p}} | x_i^{\bar{p}})\right],
\end{equation}
where the first term (reconstruction loss) represents the expected log-likelihood of the true next state under the decoder, $\mathbb{E}[\log p_\phi\left(\hat{S}_i^{t+1} | z_i^{\bar{p}}\right)]$. Maximizing this term encourages the decoder outputs $S_{i,\bar{p}}^{t+1}$ closely approximate the true next state $\hat{S}_i^{t+1}$.
Meanwhile, the second term is Kullback-Leibler (KL) divergence, which ensures the posterior distribution $q_{\xi}(z_i^{\bar{p}} \,|\, x_i^{\bar{p}})$ remains close to the prior distribution $p(z_i^{\bar{p}}|x_i^{\bar{p}})$. 
For simplicity, the prior $p(z_i^{\bar{p}} \,|\, x_i^{\bar{p}})$ is typically assumed to be a standard normal distribution, $p(z_i^{\bar{p}})=\mathcal{N}(0, I)$, independent of $x_i^{\bar{p}}$.

In the ISR module, we construct a VAE input from each phase state vector at the intersection, yielding the intermediate latent vectors $\mu_i^p$ and $\sigma_i^p$. 
We then build an intersection-specific feature vector from the latent mean vectors, $\mathbf{h}_{int} \in \mathbb{R}^{|\mathcal{P}_i| \times d_{vae}} = [\mu_i^p \;|\; p \in \mathcal{P}_i]$.
These latent vectors support supervised prediction of the next traffic state while capturing vital traffic flow dynamics.
The key insight here is to augment the VAE input with additional phase state vectors and intersection topology vectors from different agents, thereby generating diverse latent expressions that boost the Unicorn framework’s ability to represent varied traffic conditions.

\subsubsection{Contrastive Learning-Enhanced Representation}
To enhance the representation learning of latent vectors obtained through ISE, we further introduce a self-supervised contrastive learning approach. This ensures that latent variables of the same intersection are closer in the latent space, while those of different intersections are further apart, improving the distinction and expressiveness of the learned representations.

First, we define positive pairs as the latent vectors collected from the same intersection at different time steps, while negative pairs are the latent vectors collected from different intersections. 
To achieve this, we gather all latent vectors from all intersections at various time steps within a single episode and store them in a contrastive learning buffer. 
In this buffer, for a given intersection $i$, the positive pairs are defined as: ${Positive}_{i}=\left\{\left(\mu_{i}^{u}, \mu_{i}^{v}\right) \mid u \neq v\right\}$, where $\mu_{i}^{u}$ and $\mu_{i}^{v}$ represent the latent vectors of intersection $i$ at time step $u$ and $v$ derived from the previous ISE process, respectively. 
The negative pairs are defined as: ${Negative}_{i}=\left\{\left(\mu_{i}^{u}, \mu_{j}^{k}\right) \mid j \neq i, \forall k\right\}$, where $\mu_{j}^{k}$ represents the latent vectors from any other intersection $j \neq i$.
Specifically, we randomly sample batches of these pairs and employ the NT-Xent loss (Normalized Temperature-scaled Cross-Entropy loss)~\cite{wu2018unsupervised, chen2020simple, you2020graph} for contrastive learning, which is designed to measure the similarity between different learned latent vectors collected from the same agent (positive pairs) and distinguish them from different latent vectors collected from other agents (negative pairs). 
This NT-Xent loss, widely used in graph and visual representation learning, provides stable convergence and avoids costly negative sampling, making it well-suited to our network-wide ATSC task.
The NT-Xent loss for a positive sample $(u, v)$ is defined as:
\begin{equation}
\label{eqn:nt-xent}
L^{cont}_{(u,v)}\!=\!-\!\log \frac{\exp \left(\operatorname{sim}\left(\mu_i^u, \mu_i^v\right) / \tau^{cont}\right)}{\sum_j\!\sum_k\!1_{[j \neq i \; \text{or} \; k \neq u]}\! \exp\! \left(\operatorname{sim}\left(\mu_i^u, \mu_j^k\right) / \tau^{cont}\right)},
\end{equation}
where $sim(\mu_i, \mu_j)$ denotes the similarity between latent vector $\mu_i$ and $\mu_j$. 
This is typically computed using the cosine similarity $\operatorname{sim}(\mu_i, \mu_j)=\frac{\mu_i^T \mu_j}{\left\|\mu_i\right\|\left\|\mu_j\right\|}$. 
$\tau^{cont}$ is a temperature scaling parameter that helps to control the concentration level of the distribution. 
$1_{[j \neq i \text{ or } k \neq u]}$ is an indicator function that excludes self-comparisons.
By doing so, the contrastive learning-enhanced feature extraction aligns latent features within the same intersection and separates those from different intersections, thus promoting more diverse and intersection-specific feature representations for downstream RL decision-making.

\begin{figure*}[t!]
    \centering
    \includegraphics[width=0.8\linewidth]{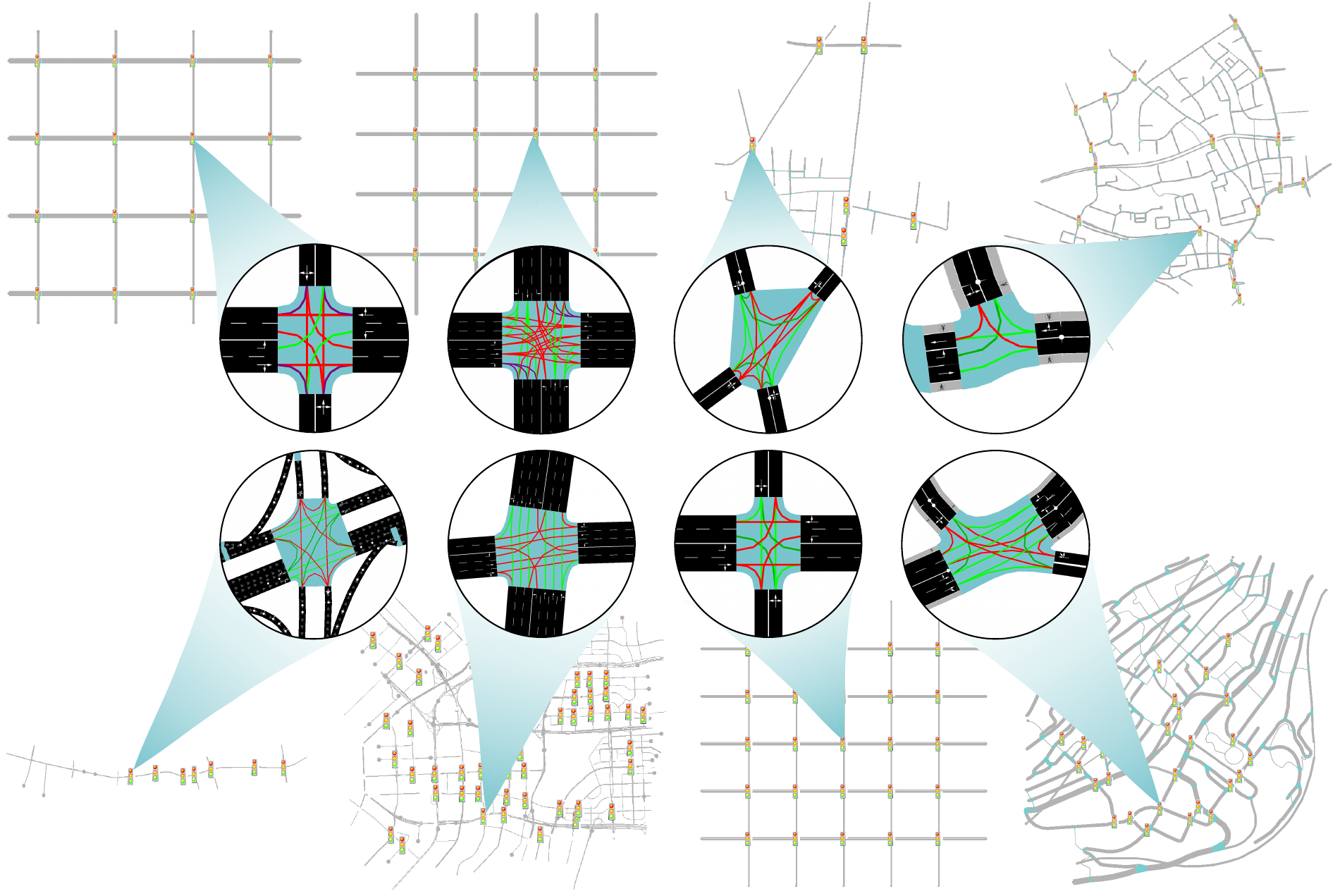}
    \centering
    \caption{
    The eight traffic datasets used for performance evaluation are displayed from left to right and top to bottom: \emph{Arterial 4$\times$4}, \emph{Grid 4$\times$4}, \emph{Cologne}, \emph{Ingolstadt}, \emph{Fenglin}, \emph{Nanshan}, \emph{Grid 5$\times$5}, and \emph{Monaco}. Typical intersection structures for each traffic network are highlighted within the circles.
    }
    \label{fig:traffic_networks}
    \vspace{-0.6cm}
\end{figure*}

\subsection{Collaborative Policy Optimization}
\label{method_cpo}
Learning efficient coordination among agents in complex, realistic traffic networks remains challenging, as agents face significant differences in their surroundings, state spaces, and action spaces.
Unlike homogeneous systems, where agents operate under similar conditions and can easily establish an implicit consensus, heterogeneous environments lack this uniformity. 
For instance, the same phase index that signals straight traffic at one intersection might indicate a left turn at another, making it difficult for agents to align their understanding of neighboring states and actions.
Building on the unified state-action representation from Sec.~\ref{method_state}, we propose a more generalizable and scalable collaborative learning algorithm. 
This algorithm enables agents to integrate neighboring states and privileged actions to capture dependencies and key interactions via an attention mechanism, while optimizing a coupled neighborhood collaborative reward to enhance coordination.

\subsubsection{Policy and Value Output}

After obtaining the aggregated feature vector $\mathbf{h}_{sp}$ (from the UTR module) and the intersection-specific feature vector $\mathbf{h}_{int}$ (from the ISR module) for all available phase state vectors, we concatenate them to form the final feature vector: $\mathbf{h} \in \mathbb{R}^{|\mathcal{P}_i| \times (d+d_{vae})} = \mathbf{Concat}(\mathbf{h}_{sp}, \mathbf{h}_{int})$. 
As shown in Fig.~\ref{fig:method}, we then compute the policy function of agent~$i$ through a linear layer (denoted as $\mathbf{f}_{\pi}$) followed by a softmax:
$\pi_i^t \in \mathbb{R}^{|\mathcal{P}_i| \times 1} = \textbf{Softmax}(\mathbf{f}_{\pi}(\mathbf{h}))$.
Furthermore, we incorporate privileged information, namely the current actions of neighboring agents, into the value function to better capture complex dynamics and interactions among agents in multi-agent environments.
This effectively reduces value estimation variance and stabilizes training by treating other agents as part of the environment~\cite{foerster2018counterfactual, kuba2021settling}.
Concretely, we feed the neighbor action vector $U_i(t)$ into a two-layer $\mathrm{MLP}_a$, yielding a high-dimensional feature vector:
$\mathbf{h_{a}}\in \mathbb{R}^{d}=\mathbf{MLP}_{a}^{(2)}(U_i)$. 
We then design an action decoder that utilizes the same multi-head cross-attention mechanism, where keys and values are obtained from $\mathbf{h_a}$, and the query is calculated from $\mathbf{h}$. The output is concatenated with $\mathbf{h}$ to produce $\mathbf{h}'$.
By adaptively capturing dependencies between phase-specific state features and neighboring action features, this design supports more accurate value estimation and fosters more effective coordination among neighboring agents.
In particular, the attention mechanism identifies the relative influence of different neighbors, further enhancing the interaction modeling essential for collaborative policy learning.
Finally, we pass $\mathbf{h}^{'}\in \mathbb{R}^{|\mathcal{P}_i| \times d}$ through a linear layer $\mathbf{f}_{v}$ to compute the state-value function:  $V_i^t \in \mathbb{R}^{1} = \textbf{Sum}(\mathbf{f}_{v}(\mathbf{h}^{'}))$.

\subsubsection{Policy Optimization}

We employ the popular RL algorithm, Proximal Policy Optimization (PPO)~\cite{schulman2017proximal} to update both the policy $\pi_{\theta}$ (parameterized by $\theta$) and the value function $V_{\Phi}$ (parameterized by $\Phi$), both of which are shared among all agents. 
Specifically, the policy loss for agent $i$ is defined as: 
\begin{equation}
L^{p}_i(\theta)= - \mathbb{E}_{\tau}\left[\min \left(\kappa_i^t(\theta) \hat{A}_i^t, \operatorname{clip}\left(\kappa_i^t(\theta), 1-\epsilon, 1+\epsilon\right) \hat{A}_i^t \right)\right], 
\end{equation}
where $\kappa(\theta)$ is the probability ratio between the current and old policies, $\hat{A}$ is the advantage estimate computed using the Generalized Advantage Estimate (GAE) approach~\cite{schulman2015high}, and $\epsilon$ defines the clipping range for the probability ratio.
Additionally, the value loss in PPO is defined as follows:
\begin{equation}
L^v_i(\Phi)=\mathbb{E}_{\tau}\left[\left(r^t_i + \gamma \, V_{\Phi, i}^{t+1} - V_{\Phi, i}^{t} \right)^2\right],
\end{equation}
which minimizes the mean-squared Temporal Difference (TD) error. 
To encourage exploration and mitigate premature convergence, we also introduce an entropy term $L_i^e(\theta)$.
By combining the VAE loss $L_i^{vae}(\theta, \Phi)$ (negative ELBO from Eq.~\ref{eqn:ELBO}) and the contrastive loss $L_i^{cont}(\theta, \Phi)$ from Eq.~\ref{eqn:nt-xent} to enhance representation learning, the overall loss for $N$ agents is defined as:
\begin{equation}
L(\theta, \Phi) = \frac{1}{N} \sum_{i=1}^{N} (L_{i}^{p} + c_1 L_{i}^{v} - c_2 L_{i}^{e} + c_3 L_i^{vae} + c_4 L_i^{cont}),
\end{equation} 
where $c_1$, $c_2$, $c_3$, and $c_4$ are constant coefficients that balance the critic loss, entropy loss, VAE loss, and contrastive loss with the policy loss during training. 
In summary, our VAE and contrastive losses act as auxiliary losses to enhance the model’s representational capacity for capturing distinctive intersection-specific features through the ISR module, whereas the RL losses remain the principal optimization objective. 
All loss terms are jointly optimized to ensure that the learned representations are well aligned with the RL objective.

\section{Experiments}
\label{experiments}
\subsection{Traffic Datasets}
To evaluate the efficacy of our proposed method, we conduct comparative experiments with a range of baseline methods across eight distinct traffic datasets using the open-source traffic simulator, SUMO~\cite{SUMO2018}.
The traffic datasets used in our experiments include \emph{Arterial 4$\times$4}, \emph{Grid 4$\times$4}, \emph{Cologne}, \emph{Ingolstadt} from RESCO~\cite{ault2021reinforcement}; \emph{Fenglin} and \emph{Nanshan} from GESA~\cite{jiang2024general}; and \emph{Grid 5$\times$5} and \emph{Monaco} from MA2C~\cite{chu2019multi}. 
The specifics of all eight traffic networks are shown in Fig.~\ref{fig:traffic_networks}.
Each dataset consists of a traffic network and its corresponding traffic flows. 
The traffic network includes structural details, such as intersection layouts, road segment configurations, and connectivity, which define its spatial organization. 
Notably, the traffic networks in \emph{Arterial 4$\times$4}, \emph{Grid 4$\times$4}, and \emph{Grid 5$\times$5} are homogeneous (i.e., identical topologies and phase settings), while those in the remaining datasets are heterogeneous.
The traffic flows specify unique movement patterns, characterized by vehicle origin-destination (OD) pairs, flow rates, and generation times, indicating how vehicles move throughout the network, with OD pairs and flow generation settings identical to the original works.
All traffic flows are time-varying, either generated synthetically with designed temporal changes or calibrated from real-world data, so that they reflect realistic and dynamic traffic conditions.
Based on the number of intersections and the average arrival rate, the eight datasets are classified as easy, medium, or hard.
Table~\ref{table:traffic_dataset} provides detailed information on each of the eight traffic datasets, including specifics of the traffic networks and traffic flows.
Detailed descriptions of the network topologies and traffic flow settings of all datasets are provided in our \href{https://www.dropbox.com/scl/fi/54bepi43mm8nkaqn54607/Supplementary_Material-T-ITS-Unicorn.pdf?rlkey=71aeqnpyg9zzyq12tlwjb1808&st=ke46mufi&dl=0}{Supplementary Material}.

\begin{table*}[t]
\centering
\caption{Intersection structure and flow details of the eight traffic datasets, where $\dagger$ denotes a homogeneous traffic network and $\ast$ denotes a heterogeneous traffic network. 
The labels (easy, medium, hard) indicate the difficulty of each dataset.}
\resizebox{0.9\linewidth}{!}{
\tiny
\begin{tabular}{c|cccc|c|cccc}
\hline
\multirow{2}{*}{Traffic Dataset
} & \multicolumn{4}{c|}{Network Structure} & \multirow{2}{*}{Volume(veh)} & \multicolumn{4}{c}{Arrival Rate(veh/min)} \\ \cline{2-5} \cline{7-10} 
 & \#Total Int & \# 2-arm & \# 3-arm & \# 4-arm &  & Mean & Std. & Max & Min \\ \hline
$\emph{Grid 4$\times$4}^{\dagger}\text{(easy)}$ & 16 & 0 & 0 & 16 & 1473.00 & 24.55 & 13.89 & 72.00 & 6.00 \\ \hline
$\emph{Arterial 4$\times$4}^{\dagger}\text{(medium)}$  & 16 & 0 & 0 & 16 & 2484.00 & 41.40 & 24.69 & 88.00 & 10.00 \\ \hline
$\emph{Cologne}^{\ast}\text{(easy)}$& 8 & 1 & 3 & 4 & 2046.00 & 34.10 & 12.65 & 77.00 & 12.00 \\ \hline
$\emph{Ingolstadt}^{\ast}\text{(hard)}$ & 21 & 0 & 17 & 4 & 4283.00 & 71.38 & 15.02 & 112.00 & 2.00 \\ \hline
$\emph{Fenglin}^{\ast}\text{(medium)}$ & 7 & 0 & 2 & 5 & 6677.00 & 111.28 & 3.75 & 117.00 & 103.00 \\ \hline
$\emph{Nanshan}^{\ast}\text{(hard)}$ & 29 & 1 & 6 & 22 & 15046.00 & 224.57 & 38.99 & 262.00 & 1.00 \\ \hline
$\emph{Grid 5$\times$5}^{\dagger}\text{(hard)}$ & 25 & 0 & 0 & 25 & 7296.00 & 121.60 & 121.94 & 752.00 & 32.00 \\ \hline
$\emph{Monaco}^{\ast}\text{(medium)}$ & 30 & 7 & 9 & 14 & 2860.00 & 47.67 & 34.29 & 151.67 & 10.83 \\ \hline
\end{tabular}
}
\vspace{-0.5cm}
\label{table:traffic_dataset}
\end{table*}

\subsection{Compared Methods}
\begin{enumerate}
    \item \textbf{Greedy}~\cite{goel2023sociallight}: Chooses the phase that releases the maximum queue length in the incoming lanes at intersections.
    \item \textbf{Max-Pressure}~\cite{varaiya2013max}: Selects the phase that minimizes intersection pressure, defined as the average difference in vehicle counts between upstream and downstream lanes.
    \item \textbf{IQL-DNN (IQLD)}~\cite{chu2019multi}: An enhanced Independent Q-learning (IQL) variant using deep neural networks (DNNs) for more accurate Q-function approximation.
    \item \textbf{IA2C}~\cite{chu2019multi}: 
    Builds on the IQL method and improves it by adopting the advantage actor-critic algorithm (A2C).
    \item \textbf{IDQN}: Similar to IQL-DNN, but utilizes convolutional layers for processing input traffic states.
    \item \textbf{IPPO}: An independent learning approach with policy updates guided by the PPO algorithm~\cite{schulman2017proximal}.
    \item \textbf{MA2C}~\cite{chu2019multi}: An advanced MARL approach that incorporates observations and fingerprints of nearby agents into the ego agent's state for stable training, and also integrates neighbors' rewards to promote agent cooperation.
    \item \textbf{FMA2C}~\cite{ma2020feudal}: A hierarchical MARL framework inspired by feudal RL, where manager agents coordinate regional behaviors and set high-level goals, while worker agents control traffic signals to achieve these goals, thereby balancing global coordination and scalability.    
    \item \textbf{CoLight}~\cite{wei2019colight}: A parameter-sharing RL method that leverages graph attention networks (GATs) to model the influences of neighbors and promote agent cooperation.
    \item \textbf{MPLight}~\cite{chen2020toward}: It is an extension of the FRAP method that adopts pressure as the state and reward, while employing the DQN algorithm for policy learning.
    \item \textbf{AttendLight}~\cite{oroojlooy2020attendlight}:
    An attention-based learning framework with parameter-sharing, designed for managing heterogeneous traffic signals, adaptable to various intersection layouts and signal phase configurations.
    \item \textbf{HeteroLight}~\cite{zhang2024HeteroLight}: A parameter-sharing approach designed specifically for heterogeneous traffic networks.
    \item \textbf{GESA}~\cite{jiang2024general}: An advanced, general ATSC method that unifies intersection topologies and employs a refined FRAP module to enable multi-scenario co-training.
\end{enumerate}
Here, \textbf{Greedy} and \textbf{Max-Pressure} are rule-based conventional ATSC baselines. The independent learning methods, including \textbf{IA2C}, \textbf{IDQN}, \textbf{IPPO}, \textbf{MA2C}, and \textbf{FMA2C}, allocate distinct parameters to each agent for independent updates. Conversely, other RL-based methods such as \textbf{MPLight}, \textbf{AttendLight}, \textbf{GESA}, and \textbf{HeteroLight} employ parameter-sharing, allowing all agents to share and update a common set of parameters.
Specifically, we apply both single training and joint training to \textbf{GESA} and our \textbf{Unicorn} method on the RESCO and GESA datasets.
Models with the single suffix (e.g., \textbf{GESA-single}, \textbf{Unicorn-single}) are trained separately on each network to focus on network-specific optimization, while models with the multiple suffix (e.g., \textbf{GESA-multiple}, \textbf{Unicorn-multiple}) are jointly trained across multiple networks (i.e., all six networks from the RESCO and GESA datasets) to learn a generalized policy through shared parameters. 
All other learning-based methods are trained separately on each traffic dataset.

\begin{table*}[]
\centering
\caption{Evaluation performance on two MA2C traffic datasets (\emph{Grid 5$\times$5} and \emph{Monaco} networks)~\cite{chu2019multi}, with the best results in bold and the second-best results underlined.}
\begin{tabular}{ccccccc}
\hline
\multicolumn{7}{c}{\emph{Grid 5$\times$5} (Hard, homogeneous network with 25 intersections)} \\ \hline
\multicolumn{1}{c|}{} & \begin{tabular}[c]{@{}c@{}}Queue Length$\downarrow$\\ (veh)\end{tabular} & \begin{tabular}[c]{@{}c@{}}Speed$\uparrow$\\ (m/sec)\end{tabular} & \begin{tabular}[c]{@{}c@{}}Intersection Delay$\downarrow$\\ (sec)\end{tabular} & \begin{tabular}[c]{@{}c@{}}Trip Completion Rate$\uparrow$\\ (veh/sec)\end{tabular} & \begin{tabular}[c]{@{}c@{}}Trip Time$\downarrow$\\ (sec)\end{tabular} & \begin{tabular}[c]{@{}c@{}}Trip Delay$\downarrow$\\ (sec)\end{tabular} \\ \hline
\multicolumn{1}{c|}{Fixed-Time} & 3.16(1.57) & 2.03(1.56) & 29.34(17.65) & 0.62(0.54) & 657.67(551.72) & 411.95(428.68) \\
\multicolumn{1}{c|}{Greedy} & 2.65(1.72) & 3.27(2.42) & 42.09(38.05) & 
\underline{0.86(0.41)} & 546.96(429.35) & 263.59(325.64) \\
\multicolumn{1}{c|}{Max-Pressure} & 2.79(1.76) & 2.93(2.19) & 39.17(37.50) & 0.82(0.42) & 572.43(404.92) & 270.41(292.18) \\ \hline
\multicolumn{1}{c|}{IQLD} & 3.51(2.24) & 2.36(2.12) & 95.92(74.70) & 0.58(0.39) & 492.71(439.87) & 285.46(368.77) \\
\multicolumn{1}{c|}{IA2C} & 3.94(2.29) & 1.52(1.09) & 81.49(48.96) & 0.40(0.26) & 760.47(510.71) & 546.28(446.36) \\
\multicolumn{1}{c|}{MA2C} & 2.83(1.49) & 2.14(1.06) & 38.44(23.92) & 0.65(0.38) & 595.30(432.10) & 383.86(346.76) \\ \hline
\multicolumn{1}{c|}{AttendLight} & 3.37(2.71) & 3.22(2.55) & 62.89(46.97) & 0.85(0.37) & 569.80(491.76) & 358.32(426.50) \\
\multicolumn{1}{c|}{HeteroLight} & \underline{1.57(1.18)} & \underline{4.55(2.29)} & \underline{22.18(16.94)} & \textbf{1.03(0.49)} & \underline{443.30(351.48)} & \underline{228.67(283.31)} \\
\multicolumn{1}{c|}{Unicorn (ours)} & \textbf{0.95(0.78)} & \textbf{5.01(2.34)} & \textbf{11.66(7.64)} & \textbf{1.03(0.64)} & \textbf{321.34(255.23)} & \textbf{137.02(197.18)}
\\ \hline
\multicolumn{7}{c}{\emph{Monaco} (Medium, heterogeneous network with 30 intersections)} \\ \hline
\multicolumn{1}{c|}{} & \begin{tabular}[c]{@{}c@{}}Queue Length$\downarrow$\\ (veh)\end{tabular} & \begin{tabular}[c]{@{}c@{}}Speed$\uparrow$\\ (m/sec)\end{tabular} & \begin{tabular}[c]{@{}c@{}}Intersection Delay$\downarrow$\\ (sec)\end{tabular} & \begin{tabular}[c]{@{}c@{}}Trip Completion Rate$\uparrow$\\ (veh/sec)\end{tabular} & \begin{tabular}[c]{@{}c@{}}Trip Time$\downarrow$\\ (sec)\end{tabular} & \begin{tabular}[c]{@{}c@{}}Trip Delay$\downarrow$\\ (sec)\end{tabular} \\ \hline
\multicolumn{1}{c|}{Fixed-Time} & 2.09(1.25) & 1.69(1.83) & \underline{66.41(53.36)} & 0.26(0.24) & 652.17(531.78) & 465.60(484.56) \\
\multicolumn{1}{c|}{Greedy} & 2.08(1.40) & 2.41(3.04) & 87.82(63.38) & 0.20(0.21) & 529.68(615.31) & 377.37(577.15) \\
\multicolumn{1}{c|}{Max-Pressure} & 2.01(1.33) & 2.42(3.01) & 85.71(61.90) & 0.21(0.21) & 545.64(608.07) & 392.80(571.18) \\ \hline
\multicolumn{1}{c|}{IQLD} & 1.97(1.16) & 0.81(1.09) & 137.34(48.19) & 0.11(0.14) & 640.61(743.76) & 520.90(714.63) \\
\multicolumn{1}{c|}{IA2C} & 1.88(1.12) & 2.14(2.76) & 86.80(48.27) & 0.22(0.20) & 482.28(555.51) & 361.73(527.91) \\
\multicolumn{1}{c|}{MA2C} & 1.54(0.78) & 1.60(1.45) & \textbf{53.34(28.88)} & 0.28(0.20) & 632.63(505.03) & 456.80(462.69) \\ \hline
\multicolumn{1}{c|}{AttendLight} & 1.46(1.05) & \underline{3.38(3.58)} & 69.41(56.94) & \textbf{0.37(0.25)} & 484.66(503.37) & 346.26(475.15) \\
\multicolumn{1}{c|}{HeteroLight} & \underline{1.22(0.76)} & \textbf{3.45(3.84)} & 102.56(73.40) & 0.34(0.23) & \underline{383.87(483.26)} & \underline{252.11(459.32)} \\
\multicolumn{1}{c|}{Unicorn (ours)} & \textbf{0.74(0.37)} & 2.97(3.21) & 112.54(67.46) & \underline{0.35(0.24)} & \textbf{341.16(489.55)} & \textbf{222.39(465.94)} 
\\ \hline
\end{tabular}
\label{table:ma2c_results}
\end{table*}

\begin{figure*}[t!]
    \centering
    \includegraphics[width=\linewidth]{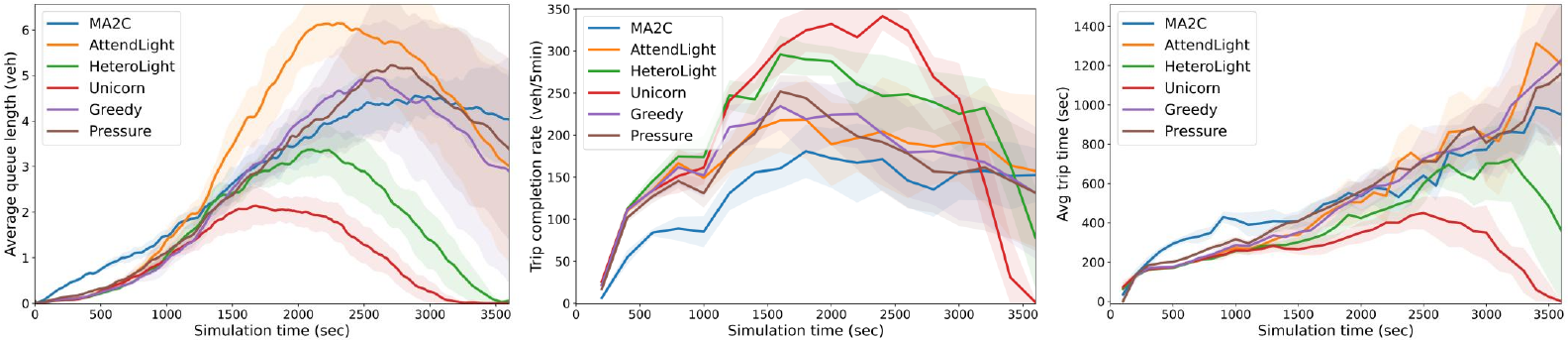}
    \caption{
    Variation of evaluation metrics (average queue length, trip completion rate, and average trip time) over simulation time (3600 seconds) on the \emph{Grid 5$\times$5} map of the MA2C dataset~\cite{chu2019multi}. 
    Here, solid lines represent the average across 10 testing episodes, with shaded areas indicating variance.
    }
    \label{fig:ma2c_results}
    \vspace{-0.5cm}
\end{figure*}

\begin{table*}[t!]
\caption{Evaluation performance on four RESCO traffic datasets (\emph{Grid 4$\times$4}, \emph{Arterial 4$\times$4}, \emph{Cologne}, and \emph{Ingolstadt} networks)~\cite{ault2021reinforcement}, where the best results are highlighted in bold and the second-best are underlined.}
\label{table:resco_results}
\centering
\begin{tabular}{ccccccc}
\hline
\multicolumn{7}{c}{\emph{Grid 4$\times$4} (Easy, homogeneous network with 16 intersections)} \\ \hline
\multicolumn{1}{c|}{} & \begin{tabular}[c]{@{}c@{}}Queue Length$\downarrow$\\ (veh)\end{tabular} & \begin{tabular}[c]{@{}c@{}}Speed$\uparrow$\\ (m/sec)\end{tabular} & \begin{tabular}[c]{@{}c@{}}Intersection Delay$\downarrow$\\ (sec)\end{tabular} & \begin{tabular}[c]{@{}c@{}}Trip Completion Rate$\uparrow$\\ (veh/sec)\end{tabular} & \begin{tabular}[c]{@{}c@{}}Trip Time$\downarrow$\\ (sec)\end{tabular} & \begin{tabular}[c]{@{}c@{}}Trip Delay$\downarrow$\\ (sec)\end{tabular} \\ \hline
\multicolumn{1}{c|}{IDQN} & 0.10(0.06) & 9.06(1.15) & 11.50(12.28) & 0.41(0.39) & 178.25(162.20) & 42.37(156.08) \\
\multicolumn{1}{c|}{IPPO} & \textbf{0.06(0.05)} & \underline{9.44(0.78)} & 1.69(0.67) & \textbf{0.42(0.36)} & \underline{164.30(56.60)} & 27.57(23.11) \\
\multicolumn{1}{c|}{FMA2C} & 0.31(0.20) & 5.53(0.81) & 21.74(5.36) & 0.41(0.36) & 284.58(151.57) & 140.95(118.08) \\ \hline
\multicolumn{1}{c|}{CoLight} & \underline{0.07(0.05)} & 9.03(0.83) & \underline{1.68(0.83)} & \textbf{0.42(0.39)} & 170.41(59.50) & 31.80(23.62) \\
\multicolumn{1}{c|}{MPLight} & 0.08(0.07) & 8.74(0.92) & 2.12(1.06) & 0.42(0.42) & 177.01(63.39) & 37.42(27.57) \\
\multicolumn{1}{c|}{GESA-single} & 0.07(0.07) & 9.29(0.96) & 2.91(2.55) & 0.42(0.40) & 169.27(64.99) & 31.61(37.31) \\
\multicolumn{1}{c|}{GESA-multiple} & \textbf{0.06(0.05)} & \textbf{9.58(0.91)} & \underline{1.41(1.00)} & 0.42(0.41) & \textbf{162.62(57.26)} & \textbf{25.37(23.28)} \\
\multicolumn{1}{c|}{Unicorn-single (ours)} & \textbf{0.06(0.05)} & 9.32(1.05) & \textbf{1.39(0.86)} & 0.40(0.43) & 165.38(58.16) & \underline{26.03(21.97)} \\
\multicolumn{1}{c|}{Unicorn-multiple (ours)} & 0.07(0.06) & 9.11(1.05) & 1.96(1.15) & 0.40(0.42) & 169.72(60.97) & 30.13(26.64) \\ \hline
\multicolumn{7}{c}{\emph{Arterial 4$\times$4} (Medium, homogeneous network with 16 intersections)} \\ \hline
\multicolumn{1}{c|}{} & \begin{tabular}[c]{@{}c@{}}Queue Length$\downarrow$\\ (veh)\end{tabular} & \begin{tabular}[c]{@{}c@{}}Speed$\uparrow$\\ (m/sec)\end{tabular} & \begin{tabular}[c]{@{}c@{}}Intersection Delay$\downarrow$\\ (sec)\end{tabular} & \begin{tabular}[c]{@{}c@{}}Trip Completion Rate$\uparrow$\\ (veh/sec)\end{tabular} & \begin{tabular}[c]{@{}c@{}}Trip Time$\downarrow$\\ (sec)\end{tabular} & \begin{tabular}[c]{@{}c@{}}Trip Delay$\downarrow$\\ (sec)\end{tabular} \\ \hline
\multicolumn{1}{c|}{IDQN} & 2.34(1.35) & 2.11(2.04) & 125.30(95.75) & 0.34(0.29) & 659.41(791.90) & 493.84(795.88) \\
\multicolumn{1}{c|}{IPPO$^{\star}$} & 4.06(2.00) & 1.07(1.87) & 253.08(166.95) & 0.11(0.16) & 1653.15(1091.73) & 1488.89(1112.30) \\
\multicolumn{1}{c|}{FMA2C$^{\star}$} & 3.31(1.50) & 1.09(0.87) & \underline{55.82(15.34)} & 0.23(0.19) & 1040.52(571.78) & 838.83(502.98) \\ \hline
\multicolumn{1}{c|}{CoLight} & 1.12(0.55) & 3.21(1.52) & \underline{29.56(15.34)} & \underline{0.53(0.31)} & 370.66(327.93) & 188.58(297.85) \\
\multicolumn{1}{c|}{MPLight} & 1.61(0.70) & 2.38(0.64) & \textbf{24.00(7.39)} & \underline{0.52(0.33)} & 465.77(237.69) & 262.53(197.70) \\
\multicolumn{1}{c|}{GESA-single} & 2.48(1.53) & 1.64(1.40) & 212.84(127.29) & 0.23(0.22) & 270.64(337.10) & 137.99(327.14) \\
\multicolumn{1}{c|}{GESA-multiple} & 1.46(0.71) & 2.85(1.65) & 114.17(74.56) & 0.45(0.25) & 415.27(454.05) & 233.72(438.79) \\
\multicolumn{1}{c|}{Unicorn-single (ours)} & \textbf{0.80(0.40)} & \textbf{4.21(1.66)} & 57.44(39.57) & \textbf{0.54(0.29)} & \underline{263.95(308.39)} & \underline{112.93(294.57)} \\ 
\multicolumn{1}{c|}{Unicorn-multiple (ours)} & \underline{0.97(0.45)} & \underline{3.72(1.65)} & 74.24(44.46) & 0.51(0.29) & \textbf{246.09(290.10)} & \textbf{98.49(276.11)} \\ \hline
\multicolumn{7}{c}{\emph{Cologne} (Easy, heterogeneous network with 8 intersections)} \\ \hline
\multicolumn{1}{c|}{} & \begin{tabular}[c]{@{}c@{}}Queue Length$\downarrow$\\ (veh)\end{tabular} & \begin{tabular}[c]{@{}c@{}}Speed$\uparrow$\\ (m/sec)\end{tabular} & \begin{tabular}[c]{@{}c@{}}Intersection Delay$\downarrow$\\ (sec)\end{tabular} & \begin{tabular}[c]{@{}c@{}}Trip Completion Rate$\uparrow$\\ (veh/sec)\end{tabular} & \begin{tabular}[c]{@{}c@{}}Trip Time$\downarrow$\\ (sec)\end{tabular} & \begin{tabular}[c]{@{}c@{}}Trip Delay$\downarrow$\\ (sec)\end{tabular} \\ \hline
\multicolumn{1}{c|}{IDQN} & 0.20(0.12) & 8.11(0.70) & 3.18(3.47) & \textbf{0.56(0.35)} & 92.99(75.53) & 11.76(53.13) \\
\multicolumn{1}{c|}{IPPO} & 0.15(0.10) & 8.35(0.67) & 0.69(0.54) & \textbf{0.56(0.35)} & \underline{89.17(52.64)} & 8.38(10.91) \\
\multicolumn{1}{c|}{FMA2C} & 1.34(0.60) & 4.50(0.76) & 17.86(6.58) & \underline{0.55(0.29)} & 186.57(166.87) & 82.72(120.48) \\ \hline
\multicolumn{1}{c|}{CoLight} & 0.22(0.13) & 7.74(0.76) & 0.88(0.49) & 0.56(0.36) & 96.65(54.96) & 12.39(13.18) \\
\multicolumn{1}{c|}{MPLight} & 0.22(0.13) & 7.81(0.69) & 0.95(0.60) & \textbf{0.56(0.34)} & 95.43(54.82) & 12.37(13.03) \\
\multicolumn{1}{c|}{GESA-single} & \underline{0.14(0.11)} & \underline{8.40(0.87)} & 2.38(3.18) & \textbf{0.56(0.35)} & 89.52(54.10) & \underline{7.71(21.54)} \\
\multicolumn{1}{c|}{GESA-multiple} & \textbf{0.12(0.11)} & \textbf{8.52(0.79)} & \textbf{0.52(0.42)} & \textbf{0.56(0.35)} & \textbf{87.96(49.85)} & \textbf{6.82(9.88)} \\
\multicolumn{1}{c|}{Unicorn-single (ours)} & \underline{0.14(0.11)} & 8.35(0.73) & \underline{0.67(0.55)} & \textbf{0.56(0.35)} & 89.98(52.22) & 8.25(11.55) \\
\multicolumn{1}{c|}{Unicorn-multiple (ours)} & 0.16(0.13) & 8.22(0.78) & 0.72(0.51) & \textbf{0.56(0.35)} & 91.94(53.39) & 9.18(12.62) \\ \hline
\multicolumn{7}{c}{\emph{Ingolstadt} (Hard, heterogeneous network with 21 intersections)} \\ \hline
\multicolumn{1}{c|}{} & \begin{tabular}[c]{@{}c@{}}Queue Length$\downarrow$\\ (veh)\end{tabular} & \begin{tabular}[c]{@{}c@{}}Speed$\uparrow$\\ (m/sec)\end{tabular} & \begin{tabular}[c]{@{}c@{}}Intersection Delay$\downarrow$\\ (sec)\end{tabular} & \begin{tabular}[c]{@{}c@{}}Trip Completion Rate$\uparrow$\\ (veh/sec)\end{tabular} & \begin{tabular}[c]{@{}c@{}}Trip Time$\downarrow$\\ (sec)\end{tabular} & \begin{tabular}[c]{@{}c@{}}Trip Delay$\downarrow$\\ (sec)\end{tabular} \\ \hline
\multicolumn{1}{c|}{IDQN} & 0.37(0.09) & 5.45(0.83) & 48.49(19.00) & 1.06(0.49) & 380.30(538.60) & 170.77(520.62) \\
\multicolumn{1}{c|}{IPPO$^{\star}$} & 1.48(0.60) & 2.65(1.95) & 108.82(43.65) & 0.62(0.33) & 931.60(937.90) & 732.29(919.47) \\
\multicolumn{1}{c|}{FMA2C} & 0.77(0.20) & 4.93(0.72) & \textbf{17.55(5.10)} & 1.05(0.46) & 396.89(275.15) & 187.06(207.61) \\ \hline
\multicolumn{1}{c|}{CoLight} & 0.45(0.18) & 5.41(1.79) & 45.20(22.64) & 0.99(0.45) & 381.20(438.26) & 194.84(410.22) \\
\multicolumn{1}{c|}{MPLight$^{\star}$} & 1.63(0.50) & 2.22(1.04) & 93.99(26.55) & 0.57(0.32) & 882.31(858.22) & 674.73(838.57) \\
\multicolumn{1}{c|}{GESA-single} & 1.15(0.43) & 2.29(1.32) & 187.24(81.44) & 0.58(0.40) & 587.41(597.63) & 363.79(552.99) \\
\multicolumn{1}{c|}{GESA-multiple} & \textbf{0.09(0.04)} & 5.96(1.27) & 61.41(45.83) & 1.02(0.58) & 283.56(228.15) & 94.93(198.87) \\
\multicolumn{1}{c|}{Unicorn-single (ours)} & 0.20(0.07) & \underline{7.17(1.47)} & \underline{21.00(14.05)} & \underline{1.08(0.47)} & \underline{252.91(194.46)} & \underline{65.84(154.93)} \\ 
\multicolumn{1}{c|}{Unicorn-multiple (ours)} & \underline{0.19(0.06)} & \textbf{7.50(0.91)} & 23.09(11.33) & \textbf{1.11(0.48)} & \textbf{247.01(210.70)} & \textbf{59.93(182.77)} \\ \hline
\end{tabular}
\vspace{-0.5cm}
\end{table*}

\subsection{Experimental Settings}

For all experiments, we maintain a total simulation time of $3600$ seconds across all datasets, adhering to the same settings as described in the baseline methods of RESCO~\cite{ault2021reinforcement}, GESA~\cite{jiang2024general}, and MA2C~\cite{chu2019multi}.
For the RESCO and GESA datasets, this includes a phase duration of 15 seconds and a yellow phase duration of 5 seconds, while for the MA2C dataset, the phase duration is 10 seconds with a yellow phase duration of 3 seconds.
Consequently, selecting the same phase results in a green light duration equal to the phase duration, while a phase change introduces the specified yellow light duration, followed by the remaining green light duration after the yellow phase.
For training, our hyperparameter configuration is as follows: a discount factor of $0.95$, a GAE factor of $0.98$, an actor learning rate of $1 \times 10^{-4}$, and a critic learning rate of $2 \times 10^{-4}$. 
We have set the scaling coefficients for value loss, entropy loss, VAE loss, and contrastive loss at $0.5$, $2 \times 10^{-3}$, $2 \times 10^{-4}$, and $1 \times 10^{-5}$, respectively. 
Additionally, the clip ratio and update epochs for the PPO algorithm are fixed at $0.2$ and $6$. 
As for contrastive learning, the temperature parameter is set to $0.2$, and the sampled batch size is fixed at $256$.
We also provide sensitivity analyses of the contrastive learning temperature and loss factor in our \href{https://www.dropbox.com/scl/fi/54bepi43mm8nkaqn54607/Supplementary_Material-T-ITS-Unicorn.pdf?rlkey=71aeqnpyg9zzyq12tlwjb1808&st=ke46mufi&dl=0}{Supplementary Material}.

In parallel, we train all learning-based baselines under the same environment settings within each dataset (e.g., simulation length per episode, phase duration, etc.), ensuring consistency across methods. 
For MA2C-based baselines\footnote{\url{https://github.com/cts198859/deeprl_signal_control}} (IQLD, IA2C, MA2C), RESCO baselines\footnote{\url{https://github.com/Pi-Star-Lab/RESCO}} (IDQN, IPPO, FMA2C, CoLight, MPLight), and the GESA baselines\footnote{\url{https://github.com/bonaldli/GESA}} (GESA-single, GESA-multiple), we follow their official open-source implementations to preserve the original training configurations and guarantee fair and reproducible evaluation. 
For other parameter-sharing methods (AttendLight, HeteroLight, and Unicorn), we use the same training setup and train until convergence, which in practice typically requires around 1500 episodes.
All experiments were conducted on an Ubuntu server with 32GB RAM, an Intel Core i9-13900KF processor (24 cores, 32 threads), and an NVIDIA RTX 4090 GPU.
In terms of computational cost during inference, the forward-pass complexity of the actor network is $\mathcal{O}(|\mathcal{P}_{max}|\,d^2)$ per agent under decentralized execution, where $|\mathcal{P}_{max}|$ is the is the maximum number of phases across intersections and $d$ is the hidden dimension.
A detailed calculation and analysis of the computational complexity are provided in our \href{https://www.dropbox.com/scl/fi/54bepi43mm8nkaqn54607/Supplementary_Material-T-ITS-Unicorn.pdf?rlkey=71aeqnpyg9zzyq12tlwjb1808&st=ke46mufi&dl=0}{Supplementary Material}.
We evaluate all methods over 10 episodes, each with a unique random seed, and ensure consistency by using identical seeds for corresponding episodes.
We adopt the evaluation metrics and their computation methods, as well as the testing settings, defined in the MA2C paper~\cite{chu2019multi}, which include average queue length (veh), average vehicle speed (m/sec), average intersection delay (sec), trip completion rate (veh/sec), average trip delay (sec), and average trip time (sec). 
Specifically, the average queue length is the number of stopped vehicles across the whole network per time step. 
The average speed is the mean vehicle speed per time step. 
The average intersection delay measures the \textbf{continuous waiting time} of vehicles (note that the waiting time resets when a vehicle starts moving again). 
The trip completion rate is the average number of vehicles reaching their destinations per step. 
The average travel time and average travel delay are the total time and delay for vehicles that successfully reach their destinations.

\subsection{Evaluation on the MA2C Dataset}

The experimental results on the \emph{Grid 5$\times$5} and \emph{Monaco} networks of the MA2C dataset, presented in Table~\ref{table:ma2c_results}, show that Unicorn achieves consistently better overall performance than traditional methods (Fixed-Time, Greedy, Max-Pressure), independent learning approaches (IQLD, IA2C, MA2C), and parameter-sharing baselines (AttendLight, HeteroLight), with superior or competitive results across most key metrics.
In the hard homogeneous \emph{Grid 5$\times$5} network, Unicorn achieves the lowest queue length of 0.95 veh and reduces both intersection delay and trip delay to 11.66 sec and 137.02 sec, respectively, representing nearly a 50\% improvement compared to HeteroLight. 
It also maintains the highest trip completion rate of 1.03 veh/sec and the lowest trip delay of 137.02 sec. 
It is worth noting that trip delay and trip time should be considered alongside the trip completion rate, as they only account for vehicles that successfully reach their destinations. 
A lower trip completion rate may lead to underestimated (lower) trip delays and times, as many vehicles still in transit are excluded.
These results highlight Unicorn’s superior capability in handling heavy traffic through learning a shared, collaborative control strategy.
In contrast, rule-based traditional ATSC methods such as Greedy and Max-Pressure perform poorly as they focus solely on optimizing immediate performance rather than long-term efficiency, resulting in high queue lengths and delays.
Independent learning methods, particularly MA2C, achieve better results than traditional approaches but remain less effective than parameter-sharing methods. 
This could be due to each agent updating its parameters independently, which introduces instability into the environment during the training process and ultimately degrades overall performance.  

Fig.~\ref{fig:ma2c_results} shows the changes in three key evaluation metrics: average queue length, trip completion rate, and trip time, over simulation time during testing on the \emph{Grid 5$\times$5} map. 
Compared to the average values in the table, these metric-time plots provide a more detailed view of how different methods perform under dynamic traffic conditions.
The results indicate that Unicorn consistently achieves better performance than the baseline methods throughout the episode, with lower queue lengths and trip delays as well as higher completion rates. 
This advantage is shown in both the low-traffic phase from 0 to 1500 seconds and the high-traffic phase from 1500 to 2500 seconds, where traffic significantly accumulates.
During the accumulation phase, although all methods show a rise and fall in queue length, Unicorn has the lowest peak and recovers more quickly, demonstrating its robustness in managing complex and dynamic traffic flows.
After 2000 seconds, its average trip time drops mainly because the traffic network has fewer remaining vehicles, with earlier congestion cleared more quickly than in other methods.
While in the medium heterogeneous \emph{Monaco} network, Unicorn continues to perform well, achieving the lowest queue length of 0.74 veh, the shortest trip delay of 222.39 sec, and the second-highest completion rate of 0.35 veh/s, outperforming AttendLight and HeteroLight across most key metrics. 
Compared to MA2C, the best independent learning method, Unicorn shows significant improvements in optimizing queue length and trip time, demonstrating the benefit of a parameter-sharing based universal framework in managing complex traffic conditions.

\begin{table*}
\caption{Evaluation performance on two GESA traffic datasets (\emph{Fenglin} and \emph{Nanshan} networks)~\cite{jiang2024general}, with the best results shown in bold and the second-best underlined.}
\label{table:gesa_results}
\centering
\begin{tabular}{ccccccc}
\hline
\multicolumn{7}{c}{\emph{Fenglin} (Medium, heterogeneous network with 7 intersections)} \\ \hline
\multicolumn{1}{c|}{} & \begin{tabular}[c]{@{}c@{}}Queue Length$\downarrow$\\ (veh)\end{tabular} & \begin{tabular}[c]{@{}c@{}}Speed$\uparrow$\\ (m/sec)\end{tabular} & \begin{tabular}[c]{@{}c@{}}Intersection Delay$\downarrow$\\ (sec)\end{tabular} & \begin{tabular}[c]{@{}c@{}}Trip Completion Rate$\uparrow$\\ (veh/sec)\end{tabular} & \begin{tabular}[c]{@{}c@{}}Trip Time$\downarrow$\\ (sec)\end{tabular} & \begin{tabular}[c]{@{}c@{}}Trip Delay$\downarrow$\\ (sec)\end{tabular} \\ \hline
\multicolumn{1}{c|}{GESA-single} & 0.77(0.28) & \underline{3.76(1.60)} & 57.88(21.72) & \underline{1.23(0.66)} & 255.55(391.00) & 118.14(289.99) \\
\multicolumn{1}{c|}{GESA-multiple} & 0.85(0.34) & \textbf{3.91(1.58)} & 57.91(22.95) & \textbf{1.24(0.64)} & 249.42(381.99) & 112.36(281.71) \\
\multicolumn{1}{c|}{Unicorn-single (ours)} & \textbf{0.37(0.08)} & 3.54(1.54) & \textbf{20.58(6.32)} & 1.20(0.64) & \underline{233.45(417.07)} & \underline{104.55(322.79)} \\ 
\multicolumn{1}{c|}{Unicorn-multiple (ours)} & \underline{0.37(0.09)} & 3.56(1.55) & \underline{20.74(6.98)} & 1.20(0.63) & \textbf{232.14(408.46)} & \textbf{102.72(312.12)} \\ \hline
\multicolumn{7}{c}{\emph{Nanshan} (Hard, heterogeneous network with 29 intersections)} \\ \hline
\multicolumn{1}{c|}{} & \begin{tabular}[c]{@{}c@{}}Queue Length$\downarrow$\\ (veh)\end{tabular} & \begin{tabular}[c]{@{}c@{}}Speed$\uparrow$\\ (m/sec)\end{tabular} & \begin{tabular}[c]{@{}c@{}}Intersection Delay$\downarrow$\\ (sec)\end{tabular} & \begin{tabular}[c]{@{}c@{}}Trip Completion Rate$\uparrow$\\ (veh/sec)\end{tabular} & \begin{tabular}[c]{@{}c@{}}Trip Time$\downarrow$\\ (sec)\end{tabular} & \begin{tabular}[c]{@{}c@{}}Trip Delay$\downarrow$\\ (sec)\end{tabular} \\ \hline
\multicolumn{1}{c|}{GESA-single} &  1.19(0.97) & 7.56(3.77) & 68.62(32.87) & 2.59(0.83) & 573.55(491.33) & 226.48(367.96)\\
\multicolumn{1}{c|}{GESA-multiple} & 0.92(0.70) & 7.68(3.59) & 65.43(30.14) & 2.68(0.86) & 578.38(488.55) & 225.38(361.85) \\
\multicolumn{1}{c|}{Unicorn-single (ours)} & \underline{0.78(0.54)} & \underline{8.27(3.38)} & \underline{31.15(14.20)} & \underline{2.85(0.91)} & \underline{540.97(453.38)} & \underline{195.04(317.21)}\\
\multicolumn{1}{c|}{Unicorn-multiple (ours)} & \textbf{0.76(0.53)} & \textbf{8.34(3.34)} & \textbf{28.77(12.48)} & \textbf{2.87(0.91)} & \textbf{537.70(445.90)} & \textbf{191.87(314.06)}\\
\hline
\end{tabular}
\vspace{-0.3cm}
\end{table*}

\subsection{Evaluation on the RESCO Dataset}
\subsubsection{Single Scenario Training Performance}
\label{sec:resco_single}
For models trained and tested on a single dataset (such as all independent learning methods, MPLight, GESA-single, and Unicorn-single), Unicorn demonstrates performance comparable to the state-of-the-art baseline GESA and slightly better than other parameter-sharing baselines, MPLight and CoLight, in both the simple homogeneous \emph{Grid 4$\times$4} network and the heterogeneous \emph{Cologne} network. 
Additionally, we observe that independent learning methods, particularly IPPO, can achieve performance similar to parameter-sharing methods in these low-complexity homogeneous and heterogeneous networks, indicating their effectiveness in simpler scenarios.
In the medium homogeneous \emph{Arterial 4$\times$4} network, Unicorn shows a clear advantage over both independent learning methods and other parameter-sharing baselines.
It reduces the queue length to 0.80 vehicles and the trip delay to 112.93 seconds, outperforming CoLight, MPLight and GESA, and demonstrating its effectiveness in handling complex and dynamic traffic conditions.
Independent learning methods, particularly IPPO and FMA2C, perform poorly on this network due to convergence issues during training, resulting in significantly lower performance. 
This highlights the inherent limitations of independent learning in more complex environments.

Finally, in the hard heterogeneous \emph{Ingolstadt} network, Unicorn achieves several-fold improvements over all baselines across all six metrics.
Compared to their performance on other networks in the RESCO dataset, parameter-sharing methods like GESA and MPLight show a clear performance decline.
This is likely due to the complex intersection topologies (e.g., varying shapes and road lengths) and dynamic traffic patterns, which pose significant challenges for parameter-sharing baselines. 
Consequently, they often converge to suboptimal solutions, limiting their ability to learn effective control strategies. 
In contrast, CoLight shows better performance than both MPLight and GESA, benefiting from its GAT-based design that enhances inter-agent cooperation.
Further analysis of this issue is provided in Sec.~\ref{sec:resco_multiple}.
In contrast, independent learning methods like IDQN and FMA2C perform better than GESA and MPLight, likely because each agent independently maintains and optimizes its own parameters.
By focusing solely on local decision-making without needing to share a unified policy across intersections, these methods are less impacted by the diverse intersection structures during training.
However, their lack of agent collaboration still limits overall performance in optimizing traffic across the entire network.

\subsubsection{Multiple Scenario Joint Training Performance}
\label{sec:resco_multiple}
For the multi-scenario joint training results, we observe a significant performance gap between GESA-single and GESA-multiple in more complex traffic networks. 
This indicates that GESA tends to converge to suboptimal solutions when trained on a single scenario, likely due to the unique intersection topology and traffic flow, which drive the shared policy learning toward local optima.
In contrast, joint training across multiple scenarios introduces more diverse and unrelated samples, improving sample diversity and reducing the risk of local optima.
This improvement is particularly evident in the \emph{Arterial 4$\times$4} and \emph{Ingolstadt} networks, where GESA-multiple outperforms GESA-single, achieving several-fold improvements in key metrics such as queue length, trip time, and trip delay.

In contrast, the performance gap between Unicorn-single and Unicorn-multiple is less significant.
On simpler networks like \emph{Grid 4$\times$4} and \emph{Cologne}, both Unicorn variants show comparable performance, whereas on more complex networks such as \emph{Arterial 4$\times$4} and \emph{Ingolstadt}, Unicorn-multiple achieves moderate improvements over Unicorn-single. 
This improvement suggests that Unicorn’s unique design, with its UTR and ISR modules, effectively captures the diverse topological and dynamic traffic features across diverse traffic networks. 
As a result, even with limited training on a single scenario, Unicorn demonstrates strong adaptability and robust performance in optimizing traffic flow. 
Nevertheless, we acknowledge that the additional sample diversity from joint training further enhances Unicorn’s adaptability, particularly in highly dynamic and complex networks. 
These findings highlight the complementary benefits of Unicorn’s inherent design and multi-scenario joint training, enabling more efficient and generalizable ATSC strategies across diverse network configurations.

Furthermore, we include a zero-shot generalization experiment, where methods co-trained on the RESCO and GESA datasets are directly transferred to the unseen \emph{Grid 5$\times$5} network. 
Unicorn achieves the best performance across all generalizable ATSC methods, showing its strong cross-scenario transferability.
We provide detailed experiment results and discussions in our \href{https://www.dropbox.com/scl/fi/54bepi43mm8nkaqn54607/Supplementary_Material-T-ITS-Unicorn.pdf?rlkey=71aeqnpyg9zzyq12tlwjb1808&st=ke46mufi&dl=0}{Supplementary Material}.

\begin{table*}[]
\centering
\caption{Ablation results for different Unicorn variants on the \emph{Grid 5$\times$5} network of MA2C dataset~\cite{chu2019multi}, with the best results in bold and the second-best results underlined.}
\begin{tabular}{ccccccc}
\hline
\multicolumn{7}{c}{\emph{Grid 5$\times$5} (Hard, homogeneous network with 25 intersections)l} \\ \hline
\multicolumn{1}{c|}{} & \begin{tabular}[c]{@{}c@{}}Queue Length$\downarrow$\\ (veh)\end{tabular} & \begin{tabular}[c]{@{}c@{}}Speed$\uparrow$\\ (m/sec)\end{tabular} & \begin{tabular}[c]{@{}c@{}}Intersection Delay$\downarrow$\\ (sec)\end{tabular} & \begin{tabular}[c]{@{}c@{}}Trip Completion Rate$\uparrow$\\ (veh/sec)\end{tabular} & \begin{tabular}[c]{@{}c@{}}Trip Time$\downarrow$\\ (sec)\end{tabular} & \begin{tabular}[c]{@{}c@{}}Trip Delay$\downarrow$\\ (sec)\end{tabular} \\ \hline
\multicolumn{1}{c|}{Unicorn (ours)} & \textbf{0.95(0.78)} & \underline{5.01(2.34)} & \textbf{11.66(7.64)} & \textbf{1.03(0.64)} & \textbf{321.34(255.23)} & \textbf{137.02(197.18)} \\ \hline
\multicolumn{1}{c|}{Unicorn w/o UTR} &  1.36(1.06) & 4.59(2.19) & 17.09(10.23) & \textbf{1.03(0.56)} & 403.43(313.62) & 196.76(242.81) \\
\multicolumn{1}{c|}{Unicorn w/o ISR} & 1.44(1.07) & 4.76(2.21) & 25.31(17.55) &  \underline{1.02(0.51)} &  396.02(322.29) & 193.92(254.69)  \\
\multicolumn{1}{c|}{Unicorn w/o Contrast} & \underline{1.07(0.89)} & \textbf{5.09(2.18)} & \underline{14.78(10.17)} & \textbf{1.03(0.59)} &  \underline{349.75(263.44)} &  \underline{155.13(208.90)}  \\
\multicolumn{1}{c|}{Unicorn w/o Collab} & 1.27(0.94) & 4.80(2.41) & 18.29(10.49) & \underline{1.02(0.55)} & 383.50(291.01) &  176.69(222.84) 
\\ \hline
\end{tabular}
\label{table:ablation_results}
\vspace{-0.3cm}
\end{table*}

\subsection{Evaluation on the GESA Dataset}
To further evaluate the performance of the proposed framework under more complex traffic scenarios, we test GESA and Unicorn on two heterogeneous networks (\emph{Fenglin} and \emph{Nanshan}) in the GESA dataset, using both single-scenario training and multi-scenario joint training setups, with results presented in Table~\ref{table:gesa_results}.
Under single-scenario training, Unicorn-single consistently outperforms GESA-single on both networks, particularly in reducing queue length and intersection delay. 
Additionally, Unicorn-single improves trip time and trip delay by 8.65\% and 11.50\%, respectively, while maintaining a comparable trip completion rate. 
Similar improvements are observed on the more complex \emph{Nanshan} network, where Unicorn-single significantly surpasses GESA-single across all metrics. 
These results highlight Unicorn's ability to identify, capture, and leverage unique topology structures and dynamic traffic features across various intersections, enabling superior strategy learning even with limited single-scenario training.

In the multi-scenario joint training setup, both methods show performance improvements, with GESA exhibiting more significant gains.
For example, on the \emph{Nanshan} network, GESA-multiple reduces queue length from 1.19 veh to 0.92 veh and intersection delay from 68.62 sec to 65.43 sec, demonstrating the effectiveness of joint training in addressing GESA’s tendency to converge to suboptimal solutions during single-scenario training. 
Despite these gains, GESA-multiple still falls short of the Unicorn variants.
Unicorn-multiple demonstrates exceptional performance on the challenging \emph{Nanshan} network, achieving a queue length of 0.76 veh, intersection delay of 28.77 sec, and trip delay of 191.87 sec, outperforming all other methods.  
Overall, while both methods benefit from joint training across multiple scenarios, Unicorn’s architecture demonstrates superior robustness and universality, achieving strong performance even under single-scenario training. 
These findings underscore the importance of Unicorn’s design, which enables more efficient ATSC strategies across diverse and complex networks.

\subsection{Ablation Study}
We conduct an ablation study to analyze the effectiveness of each key component in Unicorn by comparing it with four variants. 
The first variant, \emph{Unicorn w/o UTR}, replaces the UTR module with a simple MLP-based network to assess the efficiency of our decoder-only feature extraction network. 
The second variant, \emph{Unicorn w/o ISR}, removes the ISR module entirely to examine the importance of modeling intersection-specific topology and traffic dynamics in policy learning. 
The third variant, \emph{Unicorn w/o Contrast}, excludes the contrastive learning approach within the ISR module to evaluate its role in enhanced representation learning. 
The fourth variant, \emph{Unicorn w/o Collab}, removes collaborative learning based on neighbor state and action information to evaluate the impact of interaction-aware collaboration.
All methods are trained and tested on the \emph{Grid 5$\times$5} dataset under the same experimental settings, with results summarized in Table~\ref{table:ablation_results}.

The results show that removing the UTR module significantly increases the queue length to 1.36 veh and the intersection delay to 17.09 sec, indicating that UTR is crucial for extracting generalizable features across intersections with diverse topologies. 
Additionally, the trip time increases by around 25\%, further highlighting the importance of UTR in ensuring efficient and smooth traffic flow.
Without the ISR module, the intersection delay rises sharply to 25.31 sec, more than doubling the delay in the complete model, while the queue length increases to 1.44 veh. 
This substantial performance degradation underscores the necessity of ISR in accurately modeling localized intersection-specific topology and dynamic traffic patterns, thereby improving decision-making in complex traffic scenarios.
Furthermore, removing contrastive learning results in an increase in trip delay to 155.13 sec and trip time to 349.75 sec, indicating that contrastive learning plays a critical role in enhancing intersection-specific representations and policy diversity learning for improved overall traffic efficiency.
In the absence of collaborative learning, the queue length increases to 1.27 veh, the intersection delay rises to 18.29 sec, and the trip time extends to 383.50 sec. 
These results demonstrate the importance of collaborative learning in capturing inter-agent state-action dependencies, thus facilitating more collaborative decision-making across multiple intersections, and maintaining effective network-wide traffic optimization.
Overall, the ablation study shows the importance of each component in Unicorn, highlighting their roles in enabling universal and collaborative network-wide ATSC.

\section{Conclusion}
\label{conclusion}
In this paper, we propose Unicorn, a novel MARL framework designed for generalizable network-wide ATSC. 
We first propose a unified representation approach to map intersection states and actions from different topologies into a common structure based on traffic movements. 
To improve adaptability, we design the Universal Traffic Representation (UTR) module, which uses a decoder-only network for efficient general feature extraction.
We also introduce the Intersection Specifics Representation (ISR) module, which captures unique intersection topology and traffic dynamics using variational inference. 
To further refine the learned representations, we propose to rely on contrastive learning, enabling the model to better distinguish intersection-specific features and enhance diverse and adaptive policy learning.
Additionally, we incorporate privileged neighbor information to capture agent interactions and strengthen collaborative learning. 
Experimental results on eight traffic datasets empirically demonstrate the effectiveness and superiority of Unicorn over other ATSC methods.

An important direction for future research is to address more realistic scenarios, which may involve communication delays or disruptions that would affect information exchanges among agents, as well as temporary congestion caused by accidents, emergencies, or unexpected events.
We plan to develop robust communication mechanisms that can handle delays and failures, alongside the real-time incident adaptation strategies capable of adjusting to changing traffic conditions. 
In addition, we plan to incorporate fairness-aware mechanisms or objectives into our RL framework to mitigate any potential neglect of lower-traffic roads during policy optimization, thereby enhancing the fairness of signal control across different traffic flows.
Furthermore, we will evaluate the framework on larger-scale real-world networks to validate its scalability.
These future advances will aim to further improve the generalization and robustness of RL-based general ATSC frameworks, thus ensuring their effectiveness and reliability for deployments in dynamic, real-world scenarios.

\bibliographystyle{ieeetr}
\bibliography{sample}

@report{INRIX2023,
  title        = {INRIX 2023 Global Traffic Scorecard},
  author       = {{INRIX, Inc.}},
  year         = 2024,
  month        = jun,
  url          = {https://inrix.com/scorecard/},
  note         = {Accessed: 2025-01-10}
}

@article{kingma2013auto,
  title={Auto-encoding variational bayes},
  author={Kingma, Diederik P and Welling, Max},
  journal={arXiv preprint arXiv:1312.6114},
  year={2013}
}

@book{roess2004traffic,
  title={Traffic engineering},
  author={Roess, Royer P},
  year={2004},
  publisher={United states of Anerica}
}

@article{hunt1982scoot,
  title={The SCOOT on-line traffic signal optimisation technique},
  author={Hunt, PB and Robertson, DI and Bretherton, RD and Royle, M Cr},
  journal={Traffic Engineering \& Control},
  volume={23},
  number={4},
  year={1982}
}

@article{pr1992scats,
  title={SCATS: A Traffic Responsive Method of Controlling Urban Traffic Control/PR Lowrie},
  author={PR, Lowrie},
  journal={Roads and Traffic Authority},
  year={1992}
}

@article{varaiya2013max,
  title={Max pressure control of a network of signalized intersections},
  author={Varaiya, Pravin},
  journal={Transportation Research Part C: Emerging Technologies},
  volume={36},
  pages={177--195},
  year={2013},
  publisher={Elsevier}
}

@article{ye2015two,
  title={A two-way arterial signal coordination method with queueing process considered},
  author={Ye, Bao-Lin and Wu, Weimin and Mao, Weijie},
  journal={IEEE Transactions on Intelligent Transportation Systems},
  volume={16},
  number={6},
  pages={3440--3452},
  year={2015},
  publisher={IEEE}
}

@article{wu2018distributed,
  title={Distributed weighted balanced control of traffic signals for urban traffic congestion},
  author={Wu, Na and Li, Dewei and Xi, Yugeng},
  journal={IEEE transactions on intelligent transportation systems},
  volume={20},
  number={10},
  pages={3710--3720},
  year={2018},
  publisher={IEEE}
}

@inproceedings{xie2012schedule,
  title={Schedule-driven coordination for real-time traffic network control},
  author={Xie, Xiao-Feng and Smith, Stephen and Barlow, Gregory},
  booktitle={Proceedings of the International Conference on Automated Planning and Scheduling},
  volume={},
  pages={},
  year={2012}
}

@article{hu2019softpressure,
  title={Softpressure: A schedule-driven backpressure algorithm for coping with network congestion},
  author={Hu, Hsu-Chieh and Smith, Stephen F},
  journal={arXiv preprint arXiv:1903.02589},
  year={2019}
}

@inproceedings{goldstein2018expressive,
  title={Expressive real-time intersection scheduling},
  author={Goldstein, Rick and Smith, Stephen},
  booktitle={Proceedings of the AAAI Conference on Artificial Intelligence},
  volume={32},
  number={1},
  year={2018}
}

@article{wang2024real,
  title={Real-Time Network-Level Traffic Signal Control: An Explicit Multiagent Coordination Method},
  author={Wang, Wanyuan and Zhang, Haipeng and Qiao, Tianchi and Ma, Jinming and Jin, Jiahui and Li, Zhibin and Wu, Weiwei and Jiang, Yichuan},
  journal={IEEE Transactions on Intelligent Transportation Systems},
  year={2024},
  publisher={IEEE}
}

@article{zhu2023metavim,
  title={Metavim: Meta variationally intrinsic motivated reinforcement learning for decentralized traffic signal control},
  author={Zhu, Liwen and Peng, Peixi and Lu, Zongqing and Tian, Yonghong},
  journal={IEEE Transactions on Knowledge and Data Engineering},
  year={2023},
  publisher={IEEE}
}

@inproceedings{liu2023gplight,
  title={Gplight: grouped multi-agent reinforcement learning for large-scale traffic signal control},
  author={Liu, Yiling and Luo, Guiyang and Yuan, Quan and Li, Jinglin and Jin, Lei and Chen, Bo and Pan, Rui},
  booktitle={Proceedings of the Thirty-Second International Joint Conference on Artificial Intelligence},
  pages={199--207},
  year={2023}
}

@inproceedings{zang2020metalight,
  title={Metalight: Value-based meta-reinforcement learning for traffic signal control},
  author={Zang, Xinshi and Yao, Huaxiu and Zheng, Guanjie and Xu, Nan and Xu, Kai and Li, Zhenhui},
  booktitle={Proceedings of the AAAI Conference on Artificial Intelligence},
  volume={34},
  pages={1153--1160},
  year={2020}
}

@article{zhang2022neighborhood,
  title={Neighborhood cooperative multiagent reinforcement learning for adaptive traffic signal control in epidemic regions},
  author={Zhang, Chengwei and Tian, Yu and Zhang, Zhibin and Xue, Wanli and Xie, Xiaofei and Yang, Tianpei and Ge, Xin and Chen, Rong},
  journal={IEEE Transactions on Intelligent Transportation Systems},
  volume={23},
  number={12},
  pages={25157--25168},
  year={2022},
  publisher={IEEE}
}

@article{yang2021ihg,
  title={IHG-MA: Inductive heterogeneous graph multi-agent reinforcement learning for multi-intersection traffic signal control},
  author={Yang, Shantian and Yang, Bo and Kang, Zhongfeng and Deng, Lihui},
  journal={Neural networks},
  volume={139},
  pages={265--277},
  year={2021},
  publisher={Elsevier}
}

@article{wang2022meta,
  title={Meta-learning based spatial-temporal graph attention network for traffic signal control},
  author={Wang, Min and Wu, Libing and Li, Man and Wu, Dan and Shi, Xiaochuan and Ma, Chao},
  journal={Knowledge-based systems},
  volume={250},
  pages={109166},
  year={2022},
  publisher={Elsevier}
}

@inproceedings{ma2020feudal,
  title={Feudal multi-agent deep reinforcement learning for traffic signal control},
  author={Ma, Jinming and Wu, Feng},
  booktitle={Proceedings of the 19th international conference on autonomous agents and multiagent systems (AAMAS)},
  pages={816--824},
  year={2020}
}

@inproceedings{ruan2024coslight,
  title={Coslight: Co-optimizing collaborator selection and decision-making to enhance traffic signal control},
  author={Ruan, Jingqing and Li, Ziyue and Wei, Hua and Jiang, Haoyuan and Lu, Jiaming and Xiong, Xuantang and Mao, Hangyu and Zhao, Rui},
  booktitle={Proceedings of the 30th ACM SIGKDD Conference on Knowledge Discovery and Data Mining},
  pages={},
  year={2024}
}

@article{jiang2024x,
  title={X-Light: Cross-City Traffic Signal Control Using Transformer on Transformer as Meta Multi-Agent Reinforcement Learner},
  author={Jiang, Haoyuan and Li, Ziyue and Wei, Hua and Xiong, Xuantang and Ruan, Jingqing and Lu, Jiaming and Mao, Hangyu and Zhao, Rui},
  journal={arXiv preprint arXiv:2404.12090},
  year={2024}
}

@article{wang2024unitsa,
  title={UniTSA: A universal reinforcement learning framework for V2X traffic signal control},
  author={Wang, Maonan and Xiong, Xi and Kan, Yuheng and Xu, Chengcheng and Pun, Man-On},
  journal={IEEE Transactions on Vehicular Technology},
  year={2024},
  publisher={IEEE}
}

@inproceedings{lu2024dualight,
  title={DuaLight: Enhancing Traffic Signal Control by Leveraging Scenario-Specific and Scenario-Shared Knowledge},
  author={Lu, Jiaming and Ruan, Jingqing and Jiang, Haoyuan and Li, Ziyue and Mao, Hangyu and Zhao, Rui},
  booktitle={Proceedings of the 23rd International Conference on Autonomous Agents and Multiagent Systems},
  pages={1283--1291},
  year={2024}
}

@inproceedings{zhang2024HeteroLight,
title = {HeteroLight: A General and Efficient Learning Approach for Heterogeneous Traffic Signal Control},
author={Zhang, Yifeng and Li, Peizhuo and Fan, Mingfeng and Sartoretti, Guillaume},
booktitle={2024 IEEE/RSJ International Conference on Intelligent Robots and Systems (IROS)},
year={2024},
organization={IEEE}
}

@inproceedings{zhang2023leveraging,
  title={Leveraging queue length and attention mechanisms for enhanced traffic signal control optimization},
  author={Zhang, Liang and Xie, Shubin and Deng, Jianming},
  booktitle={Joint European Conference on Machine Learning and Knowledge Discovery in Databases},
  pages={141--156},
  year={2023},
  organization={Springer}
}

@inproceedings{zhang2022multi,
  title={Multi-agent traffic signal control via distributed RL with spatial and temporal feature extraction},
  author={Zhang, Yifeng and Damani, Mehul and Sartoretti, Guillaume},
  booktitle={International Conference on Autonomous Agents and Multiagent Systems},
  pages={106--113},
  year={2022},
  organization={Springer}
}

@book{oliehoek2016concise,
  title={A concise introduction to decentralized POMDPs},
  author={Oliehoek, Frans A and Amato, Christopher and others},
  volume={1},
  year={2016},
  publisher={Springer}
}

@inproceedings{goel2023sociallight,
  title={SocialLight: Distributed Cooperation Learning towards Network-Wide Traffic Signal Control},
  author={Goel, Harsh and Zhang, Yifeng and Damani, Mehul and Sartoretti, Guillaume},
  booktitle={Proceedings of the 2023 International Conference on Autonomous Agents and Multiagent Systems},
  pages={1551--1559},
  year={2023}
}

@article{haydari2020deep,
  title={Deep reinforcement learning for intelligent transportation systems: A survey},
  author={Haydari, Ammar and Y{\i}lmaz, Yasin},
  journal={IEEE Transactions on Intelligent Transportation Systems},
  volume={23},
  number={1},
  pages={11--32},
  year={2020},
  publisher={IEEE}
}

@article{wei2019survey,
  title={A survey on traffic signal control methods},
  author={Wei, Hua and Zheng, Guanjie and Gayah, Vikash and Li, Zhenhui},
  journal={arXiv preprint arXiv:1904.08117},
  year={2019}
}

@inproceedings{garg2018deep,
  title={Deep reinforcement learning for autonomous traffic light control},
  author={Garg, Deepeka and Chli, Maria and Vogiatzis, George},
  booktitle={2018 3rd IEEE international conference on intelligent transportation engineering (ICITE)},
  pages={214--218},
  year={2018},
  organization={IEEE}
}

@inproceedings{wei2018intellilight,
  title={Intellilight: A reinforcement learning approach for intelligent traffic light control},
  author={Wei, Hua and Zheng, Guanjie and Yao, Huaxiu and Li, Zhenhui},
  booktitle={Proceedings of the 24th ACM SIGKDD International Conference on Knowledge Discovery \& Data Mining},
  pages={2496--2505},
  year={2018}
}

@article{ma2021deep,
  title={A deep reinforcement learning approach to traffic signal control with temporal traffic pattern mining},
  author={Ma, Dongfang and Zhou, Bin and Song, Xiang and Dai, Hanwen},
  journal={IEEE Transactions on Intelligent Transportation Systems},
  volume={23},
  number={8},
  pages={11789--11800},
  year={2021},
  publisher={IEEE}
}

@inproceedings{chen2020toward,
  title={Toward a thousand lights: Decentralized deep reinforcement learning for large-scale traffic signal control},
  author={Chen, Chacha and Wei, Hua and Xu, Nan and Zheng, Guanjie and Yang, Ming and Xiong, Yuanhao and Xu, Kai and Li, Zhenhui},
  booktitle={Proceedings of the AAAI Conference on Artificial Intelligence},
  volume={34},
  pages={3414--3421},
  year={2020}
}

@article{oroojlooy2020attendlight,
  title={Attendlight: Universal attention-based reinforcement learning model for traffic signal control},
  author={Oroojlooy, Afshin and Nazari, Mohammadreza and Hajinezhad, Davood and Silva, Jorge},
  journal={Advances in Neural Information Processing Systems},
  volume={33},
  pages={4079--4090},
  year={2020}
}

@inproceedings{wei2019presslight,
  title={Presslight: Learning max pressure control to coordinate traffic signals in arterial network},
  author={Wei, Hua and Chen, Chacha and Zheng, Guanjie and Wu, Kan and Gayah, Vikash and Xu, Kai and Li, Zhenhui},
  booktitle={Proceedings of the 25th ACM SIGKDD International Conference on Knowledge Discovery \& Data Mining},
  pages={},
  year={2019}
}

@inproceedings{wei2019colight,
      title={Colight: Learning network-level cooperation for traffic signal control},
      author={Wei, Hua and Xu, Nan and Zhang, Huichu and Zheng, Guanjie and Zang, Xinshi and Chen, Chacha and Zhang, Weinan and Zhu, Yanmin and Xu, Kai and Li, Zhenhui},
      booktitle={Proceedings of the 28th ACM International Conference on Information and Knowledge Management},
      pages={1913--1922},
      year={2019}
    }

@article{wang2020stmarl,
  title={STMARL: A spatio-temporal multi-agent reinforcement learning approach for cooperative traffic light control},
  author={Wang, Yanan and Xu, Tong and Niu, Xin and Tan, Chang and Chen, Enhong and Xiong, Hui},
  journal={IEEE Transactions on Mobile Computing},
  volume={21},
  number={6},
  pages={2228--2242},
  year={2020},
  publisher={IEEE}
}

@inproceedings{zhang2022expression,
  title={Expression might be enough: Representing pressure and demand for reinforcement learning based traffic signal control},
  author={Zhang, Liang and Wu, Qiang and Shen, Jun and L{\"u}, Linyuan and Du, Bo and Wu, Jianqing},
  booktitle={International Conference on Machine Learning},
  pages={26645--26654},
  year={2022},
  organization={PMLR}
}

@inproceedings{zheng2019learning,
  title={Learning phase competition for traffic signal control},
  author={Zheng, Guanjie and Xiong, Yuanhao and Zang, Xinshi and Feng, Jie and Wei, Hua and Zhang, Huichu and Li, Yong and Xu, Kai and Li, Zhenhui},
  booktitle={Proceedings of the 28th ACM international conference on information and knowledge management},
  pages={1963--1972},
  year={2019}
}

@inproceedings{liang2022oam,
  title={OAM: An Option-Action Reinforcement Learning Framework for Universal Multi-Intersection Control},
  author={Liang, Enming and Su, Zicheng and Fang, Chilin and Zhong, Renxin},
  booktitle={Proceedings of the AAAI Conference on Artificial Intelligence},
  volume={},
  pages={},
  year={2022}
}

@article{mannion2016experimental,
  title={An experimental review of reinforcement learning algorithms for adaptive traffic signal control},
  author={Mannion, Patrick and Duggan, Jim and Howley, Enda},
  journal={Autonomic road transport support systems},
  pages={47--66},
  year={2016},
  publisher={Springer}
}

@article{chu2019multi,
  title={Multi-agent deep reinforcement learning for large-scale traffic signal control},
  author={Chu, Tianshu and Wang, Jie and Codec{\`a}, Lara and Li, Zhaojian},
  journal={IEEE Transactions on Intelligent Transportation Systems},
  volume={21},
  number={3},
  pages={1086--1095},
  year={2019},
  publisher={IEEE}
}

@article{chung2014empirical,
  title={Empirical evaluation of gated recurrent neural networks on sequence modeling},
  author={Chung, Junyoung and Gulcehre, Caglar and Cho, KyungHyun and Bengio, Yoshua},
  journal={arXiv preprint arXiv:1412.3555},
  year={2014}
}

@article{vaswani2017attention,
  title={Attention is all you need},
  author={Vaswani, Ashish and Shazeer, Noam and Parmar, Niki and Uszkoreit, Jakob and Jones, Llion and Gomez, Aidan N and Kaiser, {\L}ukasz and Polosukhin, Illia},
  journal={Advances in neural information processing systems},
  volume={30},
  year={2017}
}

@article{schulman2017proximal,
  title={Proximal policy optimization algorithms},
  author={Schulman, John and Wolski, Filip and Dhariwal, Prafulla and Radford, Alec and Klimov, Oleg},
  journal={arXiv preprint arXiv:1707.06347},
  year={2017}
}

@article{schulman2015high,
  title={High-dimensional continuous control using generalized advantage estimation},
  author={Schulman, John and Moritz, Philipp and Levine, Sergey and Jordan, Michael and Abbeel, Pieter},
  journal={arXiv preprint arXiv:1506.02438},
  year={2015}
}

@inproceedings{SUMO2018,
          title = {Microscopic Traffic Simulation using SUMO},
         author = {Pablo Alvarez Lopez and Michael Behrisch and Laura Bieker-Walz and Jakob Erdmann and Yun-Pang Fl{\"o}tter{\"o}d and Robert Hilbrich and Leonhard L{\"u}cken and Johannes Rummel and Peter Wagner and Evamarie Wie{\ss}ner},
      publisher = {IEEE},
      booktitle = {The 21st IEEE International Conference on Intelligent Transportation Systems},
           year = {2018},
        journal = {IEEE Intelligent Transportation Systems Conference (ITSC)},
            url = {https://elib.dlr.de/124092/}
 }

@inproceedings{ault2021reinforcement,
  title={Reinforcement learning benchmarks for traffic signal control},
  author={Ault, James and Sharon, Guni},
  booktitle={Thirty-fifth Conference on Neural Information Processing Systems Datasets and Benchmarks Track (Round 1)},
  year={2021}
}

@article{jiang2024general,
  title={A general scenario-agnostic reinforcement learning for traffic signal control},
  author={Jiang, Haoyuan and Li, Ziyue and Li, Zhishuai and Bai, Lei and Mao, Hangyu and Ketter, Wolfgang and Zhao, Rui},
  journal={IEEE Transactions on Intelligent Transportation Systems},
  year={2024},
  publisher={IEEE}
}

@inproceedings{chen2020simple,
  title={A simple framework for contrastive learning of visual representations},
  author={Chen, Ting and Kornblith, Simon and Norouzi, Mohammad and Hinton, Geoffrey},
  booktitle={International conference on machine learning},
  pages={1597--1607},
  year={2020},
  organization={PMLR}
}

@inproceedings{foerster2018counterfactual,
  title={Counterfactual multi-agent policy gradients},
  author={Foerster, Jakob and Farquhar, Gregory and Afouras, Triantafyllos and Nardelli, Nantas and Whiteson, Shimon},
  booktitle={Proceedings of the AAAI conference on artificial intelligence},
  volume={32},
  number={1},
  year={2018}
}

@article{kuba2021settling,
  title={Settling the variance of multi-agent policy gradients},
  author={Kuba, Jakub Grudzien and Wen, Muning and Meng, Linghui and Zhang, Haifeng and Mguni, David and Wang, Jun and Yang, Yaodong and others},
  journal={Advances in Neural Information Processing Systems},
  volume={},
  pages={},
  year={2021}
}

@article{zhou2024cooperative,
  title={Cooperative Traffic Signal Control Using a Distributed Agent-Based Deep Reinforcement Learning With Incentive Communication},
  author={Zhou, Bin and Zhou, Qishen and Hu, Simon and Ma, Dongfang and Jin, Sheng and Lee, Der-Horng},
  journal={IEEE Transactions on Intelligent Transportation Systems},
  year={2024},
  publisher={IEEE}
}

@article{lin2023temporal,
  title={Temporal Difference-Aware Graph Convolutional Reinforcement Learning for Multi-Intersection Traffic Signal Control},
  author={Lin, Wei-Yu and Song, Yun-Zhu and Ruan, Bo-Kai and Shuai, Hong-Han and Shen, Chih-Ya and Wang, Li-Chun and Li, Yung-Hui},
  journal={IEEE Transactions on Intelligent Transportation Systems},
  year={2023},
  publisher={IEEE}
}

@article{chen2024learning,
  title={Learning Multi-Intersection Traffic Signal Control via Coevolutionary Multi-Agent Reinforcement Learning},
  author={Chen, Wubing and Yang, Shangdong and Li, Wenbin and Hu, Yujing and Liu, Xiao and Gao, Yang},
  journal={IEEE Transactions on Intelligent Transportation Systems},
  year={2024},
  publisher={IEEE}
}

@article{zhangCoordLight,
  title={CoordLight: Learning Decentralized Coordination for Network-Wide Traffic Signal Control},
  author={Zhang, Yifeng and Goel, Harsh and Li, Peizhuo and Damani, Mehul and Chinchali Sandeep and Sartoretti, Guillaume},
  journal={Accepted by IEEE Transactions on Intelligent Transportation Systems},
  year={2025},
  publisher={IEEE}
}

@article{you2020graph,
  title={Graph contrastive learning with augmentations},
  author={You, Yuning and Chen, Tianlong and Sui, Yongduo and Chen, Ting and Wang, Zhangyang and Shen, Yang},
  journal={Advances in neural information processing systems},
  volume={33},
  pages={5812--5823},
  year={2020}
}

@inproceedings{wu2018unsupervised,
  title={Unsupervised feature learning via non-parametric instance discrimination},
  author={Wu, Zhirong and Xiong, Yuanjun and Yu, Stella X and Lin, Dahua},
  booktitle={Proceedings of the IEEE conference on computer vision and pattern recognition},
  pages={},
  year={2018}
}

@article{wang2024traffic,
  title={Traffic signal cycle control with centralized critic and decentralized actors under varying intervention frequencies},
  author={Wang, Maonan and Chen, Yirong and Kan, Yuheng and Xu, Chengcheng and Lepech, Michael and Pun, Man-On and Xiong, Xi},
  journal={IEEE Transactions on Intelligent Transportation Systems},
  year={2024},
  publisher={IEEE}
}

\end{document}